\pdfoutput=1

\documentclass[11pt]{article}

\usepackage{acl}

\usepackage{times}
\usepackage{latexsym}

\usepackage[T1]{fontenc}

\usepackage[utf8]{inputenc}

\usepackage{microtype}

\usepackage{inconsolata}

\usepackage{graphicx}
\usepackage{array}
\usepackage{booktabs}
\usepackage{multirow}
\usepackage{makecell}
\usepackage{amsmath, amssymb}
\usepackage{todonotes}
\graphicspath{{Figures/}}
\title{Language-specific Neurons Do Not Facilitate Cross-Lingual Transfer}

\author{
 \textbf{Soumen Kumar Mondal\textsuperscript{$\ddagger$}},
 \textbf{Sayambhu Sen\textsuperscript{$\S$}},
 \textbf{Abhishek Singhania\textsuperscript{$\S$}},
 \\
 \textbf{Preethi Jyothi\textsuperscript{$\ddagger$}}
\\
\\
 \textsuperscript{$\ddagger$}
Indian Institute of Technology Bombay, India,\\
 \textsuperscript{$\S$}Amazon Alexa 
 \\
\texttt{\normalsize{\{soumenkm,pjyothi\}@iitb.ac.in, \{sensayam, mrabhsin\}@amazon.com}
 }
}

\begin{document}
	\maketitle
	\begin{abstract}
    Multilingual large language models (LLMs) aim towards robust natural language understanding across diverse languages, yet their performance significantly degrades on low-resource languages. This work explores whether existing techniques to identify language-specific neurons can be leveraged to enhance cross-lingual task performance of low-resource languages. We conduct detailed experiments covering existing language-specific neuron identification techniques (such as Language Activation Probability Entropy and activation probability-based thresholding) and neuron-specific LoRA fine-tuning with models like Llama 3.1 and Mistral Nemo. We find that such neuron-specific interventions are insufficient to yield cross-lingual improvements on downstream tasks (XNLI, XQuAD) in low-resource languages.
    This study highlights the challenges in achieving cross-lingual generalization and provides critical insights for multilingual LLMs%
    \footnote{Code is available at GitHub: \url{https://github.com/csalt-research/LangSpecificNeurons}}.
	\end{abstract}
	
	\section{Introduction}
	
	Acquiring multilingual capabilities in LLMs remains a challenge, particularly for low-resource languages \cite{  4hangya-etal-2022-improving, 2conneau-etal-2020-unsupervised, 1lample2019crosslingual}. Despite their remarkable success in tasks that require cross-lingual transfer, models such as Llama 3.1 \cite{5grattafiori2024llama3herdmodels}  and Mistral Nemo \cite{6mistral-nemo} do not perform consistently across languages, particularly underperforming on low-resource languages~\cite{8touvron2023llamaopenefficientfoundation, 7hu2020xtrememassivelymultilingualmultitask}. 
    This is largely due to the imbalance in high-quality training data across languages, thus limiting the ability of multilingual models to effectively scale to low-resource languages \cite{8touvron2023llamaopenefficientfoundation, 3xue-etal-2021-mt5}. 
    
	A tool that has recently emerged to better understand the nature of multilinguality in these LLMs is the use of \emph{language-specific neurons}~\cite{11duan-etal-2025-unveiling, tang-etal-2024-language, 10zhang2024multilingualknowledgeeditinglanguageagnostic}. These neurons are claimed to encode unique language-specific features pertaining to each language, thus potentially enabling targeted language interventions. Previous studies \cite{12kojima-etal-2024-multilingual, 13zhao2024tracingrootsfactsmultilingual, tang-etal-2024-language} have demonstrated that these neurons play an important role in language generation tasks. However, the extent to which these neurons contribute to or affect cross-lingual transfer to low-resource languages when evaluated on downstream tasks such as natural language inference (XNLI) and question answering (XQuAD) remains unclear.
	
	In this study, we systematically probe the role of language-specific neurons in facilitating cross-lingual transfer within multilingual LLMs. By utilizing existing techniques to identify language-specific neurons such as Language Activation Probability Entropy (LAPE)~\cite{tang-etal-2024-language} and Low Rank Adaptation (LoRA)-based fine-tuning \cite{14hu2021loralowrankadaptationlarge}, we aim to identify and analyze neurons that mainly contribute towards language-specific representations. Our experiments span two popular cross-lingual benchmarks, XNLI for NLI~\cite{15conneau2018xnlievaluatingcrosslingualsentence} and XQuAD for QA~\cite{16Artetxe_2020}. After identifying language-specific neurons using existing techniques for a target language, we modify the activations of these language-specific neurons using different aggregation schemes in an attempt to amplify their role in cross-lingual transfer. 
	
	Our results show that such test-time (training-free) interventions via language-specific neurons are not very effective in enabling cross-lingual transfer, yielding very modest overall performance improvements of less than 1 absolute point in accuracy for low-resource languages. Fine-tuning strategies like neuron freezing and activation substitution were shown to significantly impact generation \cite{lai-etal-2024-style, 12kojima-etal-2024-multilingual} but do not show any consistent impact on cross-lingual task performance. A deeper analysis revealed that language-specific neurons often lack independence and we hypothesize that this polysemantic nature of neuron activations limits the effectiveness of targeted adjustments in multilingual LLMs \cite{17elhage2022superposition}. 
	
	\section{Methodology}
	The goal of this work is to explore whether targeting language-specific neurons in multilingual LLMs can be used to improve downstream performance on tasks such as XNLI and XQuAD. Previous studies \cite{22zhao2024largelanguagemodelshandle, 12kojima-etal-2024-multilingual, tang-etal-2024-language, 23duan-etal-2025-unveiling} have shown that distinct neuron subsets exist in multilingual models that encode language-specific features. 
    Prior work \cite{24bhattacharya2023unveilingmultilingualitytransformermodels} further indicates that language-specific representations are largely prevalent within feedforward networks.
    
    While prior work focused on how deactivating language-specific neurons degrades the quality of language generation, there has been little investigation into whether activating or fine-tuning these neurons can positively influence task performance \cite{18zhao2024largelanguagemodelshandle, lai-etal-2024-style}. This forms the main motivation for our work. We aim to evaluate the role of language-specific neurons by aiming to enhance cross-lingual task performance through targeted neuron interventions. Our results indicate that manipulating language-specific neurons, either by activating or fine-tuning them, does not lead to significant improvements in downstream task performance.
	
	\subsection{Language-Specific Neuron Identification}
	In LLMs, a \textit{neuron} corresponds to the output of the non-linear activation function within a feedforward layer. Let \(L\) be the total number of feedforward layers and \(d_f\) be the dimensionality of each feedforward layer. Each neuron is uniquely identified by a pair of indices \((i,j)\), where \( i \in \{1,2,\dots,L\}\) denotes the layer index and \(j \in \{1,2,\dots,d_f\}\) denotes the position within the hidden dimension of the feedforward network. As our main approach, we employ the LAPE method \cite{tang-etal-2024-language} to identify language-specific neurons. For a given language \(l\) and a neuron indexed by \((i, j)\), let \(h_{i,j}^l(x)\) denote the activation of that neuron for an input sentence \(s\). We define the activation probability of this neuron as:
    \[
    \mathbb{P}\left(h_{i,j}^l(s) > 0\right) := \mathbb{E}_{s \sim D_l}\left[ \mathbb{I}( h_{i,j}^l(s) > 0 ) \right],
    \]
    where \( D_l \) represents the corpus in language \( l \) and \(\mathbb{I}(\cdot)\) is the indicator function that equals 1 if the condition is satisfied and 0 otherwise. Formally, the LAPE score for a neuron \((i, j)\) is defined as:
    \begin{align*}
    \text{LAPE}(i, j) &= - \sum_{l=1}^k P_{i,j}^l \log P_{i,j}^l, \nonumber \\
    P_{i,j}^l &= \frac{\mathbb{P}\left(h_{i,j}^l(x) > 0\right) }{\sum_{l' \in \mathcal{L}} \mathbb{P}\left(h_{i,j}^{l'}(x) > 0\right)} \nonumber
    \end{align*}
    where \(P_{i,j}^l\) represents the normalized activation probability of neuron \((i, j)\) for language \(l\), and \(k\) denotes the total number of languages in the set $\mathcal{L}$. Neurons with low LAPE values are deemed to be language-specific since they exhibit high activation probabilities for only a limited subset of languages. We note here that the LAPE method is dependent on the choice of the language set $\mathcal{L}$ used for calculating the activation probability distributions. To address this limitation, we propose a simple alternative that does not have such a dependency.

	Existing methods \cite{tang-etal-2024-language, 19xie2021importancebasedneuronallocationmultilingual} often consider neurons to be relevant to a language if their activation is greater than 0, and quantify this as a relevance score computed as $r_{i,j}^l = \mathbb{E}[\mathbb{I}(h_{i,j}^l > 0)]$ where \(h_{i,j}^l\) is the activation of neuron \((i, j)\) for language \(l\). However, it overlooks the possibility that negative activations can also carry meaningful information. To account for this, we propose an activation statistics-based approach. Instead of relying on a threshold of $0$, we consider neurons as relevant if their activation exceeds a chosen percentile threshold of the overall activation distribution. For example, the relevance of a neuron based on the 90th percentile is defined as $r_{i,j}^l = \mathbb{E}[\mathbb{I}(h_{i,j}^l > P_{90}(h_{i,j}^l))]$ where \(P_{90}(h_{i,j}^l)\) is the 90th percentile of the activation values for neuron \((i, j)\) in language \(l\). We call this technique \emph{Activation Probability 90p} which is entirely based on neuron activations and avoids the language set dependency issue inherent to LAPE. Neurons are then ranked based on their relevance scores, and the top \(m\) neurons are selected as being language-specific. More details on language neuron identification can be found in Appendix \ref{app:A}. 
	
	\subsection{Neuron Fine-Tuning using LoRA}\label{sec:lora}
    LoRA~\cite{14hu2021loralowrankadaptationlarge} is employed to efficiently fine-tune only the neurons identified as language-specific in the MLP layers. Let \(\mathbf{W} \in \mathbb{R}^{d \times k}\) be a pre-trained weight matrix; LoRA adds a trainable update \(\Delta \mathbf{W}\) such that $\mathbf{W}' = \mathbf{W} + \Delta \mathbf{W}$. To remain parameter-efficient, \(\Delta \mathbf{W}\) is factorized into two low-rank matrices \(\mathbf{B} \in \mathbb{R}^{d \times r}\) and \(\mathbf{A} \in \mathbb{R}^{r \times k}\): $\Delta \mathbf{W} = \mathbf{B}\mathbf{A}, 
    \;
    r \ll \min(d,\,k).$ In the forward pass, the feedforward layer computes
    \[
    \mathbf{y} = \left(\mathbf{W} + \Delta \mathbf{W}\right)\mathbf{x} = \mathbf{W}\mathbf{x} + \mathbf{B}\mathbf{A}\mathbf{x},
    \]
    with only \(\mathbf{B}\) and \(\mathbf{A}\) being trainable. To restrict updates to language-specific neurons, we define a binary mask \(\mathbf{M} \in \{0,1\}^{d \times k}\). If \(\mathbf{M}_{i,j} = 1\), the \(j\)-th neuron of layer \(i\) is considered language-specific and thus it will be trained; otherwise, it will remain frozen. The effective LoRA update thus becomes:
    \[
    \Delta \mathbf{W} \leftarrow \mathbf{M}\otimes (\mathbf{B}\mathbf{A}),
    \]
    where \(\otimes\) denotes an element-wise multiplication. 
    Therefore, $\Delta \mathbf{W}_{i,j} = 0$ if $ \mathbf{M}_{i,j} = 0.$ Hence, the forward pass is given by:
    \[
    \mathbf{y} = \left(\mathbf{W} + \mathbf{M} \otimes (\mathbf{B}\mathbf{A})\right)\mathbf{x},
    \]
    and only those sub-blocks of \(\mathbf{B}\) and \(\mathbf{A}\) associated with masked entries of \(1\) are trainable. In addition to these masked LoRA updates, the classification head and attention layers are fine-tuned to maintain overall task performance, while all remaining parameters (including \(\mathbf{W}\) itself) remain frozen.
	
	\section{Experimental Setup}
	\subsection{Datasets, Tasks, and Models}
	To identify language-specific neurons, we use a subset of the Wikipedia \cite{20wikidump} dataset spanning 16 languages: \emph{en, fr, es, vi, id, ja, zh, bn, hi, ta, te, mr, ur, kn, ml, pa} \footnote{\url{https://en.wikipedia.org/wiki/List_of_ISO_639_language_codes}}. However, only a subset of these languages will be used for evaluation as mentioned in Section \ref{sec:results}. The dataset creation process is outlined in Appendix \ref{app:A}. For fine-tuning experiments aimed at evaluating task performance, we use two popular multilingual benchmarks: the XNLI dataset \cite{15conneau2018xnlievaluatingcrosslingualsentence} for NLI and the XQuAD dataset \cite{16Artetxe_2020} for QA. In our experiments, we use two pretrained LLMs: Llama 3.1 (8B) \cite{5grattafiori2024llama3herdmodels} and Mistral Nemo (12B) \cite{6mistral-nemo}. More details such as tasks, models, optimizer and hyper-parameters used in LoRA can be found in Appendix \ref{app:B}.
	
	\subsection{Experiment Design}\label{sec:exp}
	The primary goal of this work is to improve the zero-shot performance of the model on target languages, without using target language training data. 
    
    \noindent \textbf{Zero-Shot Transfer.} The model is fine-tuned on task-specific data from a source language and evaluated on task-specific test data for a target language. We assume access only to task-specific training data in the source language, and no target language task-specific data. Our goal is to improve over zero-shot transfer using (1) test-time language-specific neuron intervention and (2) language-specific neuron fine-tuning, detailed below.
	
	\noindent \textbf{(1) Test-time Neuron Intervention.} We train the LLM on the task-specific training dataset in the source language and evaluate its performance on the task-specific test dataset in the target language. During evaluation, we modify the activations of the target language neurons in the forward pass using a range of statistical aggregates computed based on the Wikipedia dataset of target languages. 
	
	\noindent \textbf{(2) Language Neuron Fine-Tuning.} We fine-tune the language-specific neurons as detailed in Section \ref{sec:lora}. We explore three different setups for fine-tuning: (a) Fine-tuning only the source language-specific neurons, (b) Fine-tuning only the target language-specific neurons, (c) Fine-tuning both the source and target language-specific neurons. After fine-tuning, we evaluate the model by performing test-time interventions on the target language-specific neurons.
	
	\section{Results and Analysis}\label{sec:results}
    
    \subsection{Zero Shot Transfer Performance}
	In all our experiments, we use English (\textit{en}) as the source language. For the XNLI task, we evaluate the model’s zero-shot performance on Vietnamese (\textit{vi}), Hindi (\textit{hi}), and Urdu (\textit{ur}), while for the XQuAD task, we consider Vietnamese (\textit{vi}), Hindi (\textit{hi}), and Chinese (\textit{zh}) as target languages. These target languages are selected due to their relatively lower performance in the XNLI and XQuAD benchmark results \cite{16Artetxe_2020, 15conneau2018xnlievaluatingcrosslingualsentence}, making them strong candidates for evaluating improvements in cross-lingual transfer. For the XNLI task, we use a subset of 100,000 training samples, which corresponds to 25\% of the full training dataset \cite{21asai-etal-2024-buffet}. For the XQuAD task, we utilize the entire training dataset. Table \ref{tab:xnli} and Table \ref{tab:xquad} present the zero-shot results for both tasks.

    \begin{table}[ht!]
		\centering
		 \renewcommand{\arraystretch}{1.15} 
		\setlength{\tabcolsep}{6pt} 
		\begin{small}
			\begin{tabular}{|c|c|c|c|c|c|}
				\hline
				 \textbf{EL} & \textbf{No Int} & \textbf{Int} $\mathbf{\mu}$ &  \textbf{Int P90} & \textbf{Int 0} & \textbf{Int P10} \\ 
				\hline
				\multicolumn{6}{|c|}{\textit{Llama 3.1 with LAPE}} \\ 
				\hline
				vi & \textbf{80.5} & 79.5 & 79.0 & 79.8 & 77.7 \\ 
				hi & 75.0 &  \textbf{75.2} & 74.9 & 74.4 & 75.1 \\ 
				ur & 70.0 & \textbf{70.4} & 69.3 & 68.5 & 68.7 \\ 
				\hline
				\multicolumn{6}{|c|}{\textit{Llama 3.1 with Act Prob 90p}} \\ 
				\hline
				vi & \textbf{80.5} & 78.2 & 79.3 & 79.0 & 77.4 \\ 
				hi & \textbf{75.0} & 74.1 & 71.8 & 73.7 & 74.6 \\ 
				ur & \textbf{70.0} & 69.7 & 69.3 & 69.6 & 69.5 \\ 
				\hline
				\multicolumn{6}{|c|}{\textit{Mistral Nemo with LAPE}} \\ 
				\hline
				vi & 80.5 & 80.4 & \textbf{80.6} & 79.2 & 80.5 \\ 
				hi & \textbf{76.1} & 69.8 & 66.9 & 74.9 & 72.4 \\ 
				ur & 66.8 & 66.5 & 67.0 & \textbf{66.9} & 65.4 \\ 
				\hline
				\multicolumn{6}{|c|}{\textit{Mistral Nemo with Act Prob 90p}} \\ 
				\hline
				vi & 80.5 & 67.4 & \textbf{81.1} & 79.8 & 40.7 \\ 
				hi & \textbf{76.1} & 72.2 & 74.5 & 74.5 & 66.3 \\ 
				ur & \textbf{66.8} & 65.9 & 61.3 & 66.4 & 61.6 \\ 
				\hline
			\end{tabular}
		\end{small}
		\caption{\footnotesize XNLI performance across different models and intervention methods. "No Int" represents zero-shot performance without intervention, while "Int $\mu$", "Int P90", "Int 0", and "Int P10" denote test-time interventions using mean, 90th percentile, zero, and 10th percentile activations, respectively. The best performance for each evaluation language (EL) is highlighted in bold.}
		\label{tab:xnli}
	\end{table}

    \begin{table}[ht!]
		\centering
		 \renewcommand{\arraystretch}{1.15} 
		\setlength{\tabcolsep}{3.25pt} 
		\begin{small}
		\begin{tabular}{|c|c|c|c|c|c|}
				\hline
				\textbf{EL} & \textbf{No Int} & \textbf{Int} $\mathbf{\mu}$ &  \textbf{Int P90} & \textbf{Int 0} & \textbf{Int P10} \\ 
				\hline
				\multicolumn{6}{|c|}{\textit{Llama 3.1 with LAPE}} \\ 
				\hline
				vi & \textbf{41 (73.5)} & 40 (72.9) & 31 (69.5) & 32 (69.2) & 10 (43.2) \\ 
				hi & 38 (64.1) & \textbf{40 (65.5)} & 36 (65.4) & 23 (49.9) & 37 (62.8) \\ 
				zh & \textbf{56 (77.5)} & 10 (62.8) & 3 (56.1) & 33 (63.2) & 33 (63.2) \\ 
				\hline
				\multicolumn{6}{|c|}{\textit{Llama 3.1 with Act Prob 90p}} \\ 
				\hline
				vi & 41 (73.6) & 39 (73.0) & 23 (64.9) & \textbf{42 (73.8)} & 36 (70.3) \\ 
				hi & \textbf{38 (64.1)} & 34 (60.7) & 36 (62.8) & 38 (62.9) & 31 (58.6) \\ 
				zh & 56 (77.5) & \textbf{61 (80.7)} & 56 (78.8) & 55 (78.5) & 50 (73.6) \\ 
				\hline
				\multicolumn{6}{|c|}{\textit{Mistral Nemo with LAPE}} \\ 
				\hline
				vi & 39 (74.6) & \textbf{42 (76.8)} & 40 (75.0) & 13 (45.0) & 11 (41.2) \\ 
				hi & \textbf{38 (66.9)} & 35 (65.9) & 37 (66.6) & 22 (51.7) & 36 (66.1) \\ 
				zh & \textbf{47 (74.9)} & 24 (74.0) & 0 (61.6) & 14 (53.3) & 24 (68.9) \\ 
				\hline
				\multicolumn{6}{|c|}{\textit{Mistral Nemo with Act Prob 90p}} \\ 
				\hline
				vi & \textbf{39 (74.6)} & 11 (43.3) & 29 (63.9) & 39 (74.5) & 0 (6.5) \\ 
				hi & \textbf{38 (66.9)} & 26 (54.4) & 37 (68.9) & 33 (63.8) & 0 (11.9) \\ 
				zh & 47 (74.9) & 46 (77.4) & 20 (59.8) & \textbf{48 (76.2)} & 0 (17.0) \\ 
				\hline
			\end{tabular}
		\end{small}
		\caption{\footnotesize XQuAD performance across different models and intervention methods. The intervention strategies are the same as described in Table~\ref{tab:xnli}. The values indicate Exact Match (EM) scores, with F1 scores in parentheses.}
		\label{tab:xquad}
	\end{table}
	
	\subsection{Impact of Test-Time Intervention}
	
	\paragraph{Test-Time Interventions Do Not Improve Performance.}
    Tables \ref{tab:xnli} and \ref{tab:xquad} show that test-time interventions fail to consistently improve zero-shot transfer performance. Instead, they often disrupt the task-specific information encoded in the activations. This suggests that language-specific neurons in LLMs are not purely language-dependent but also contribute to task-relevant computations. Overwriting their activations with statistical values removes essential information required for solving the task due to the polysemantic nature of neuron activations \cite{17elhage2022superposition}. We also experiment with different approaches for identifying language neurons, including LAPE and activation probability-based methods (e.g., 90th percentile); no significant improvements are observed. From the results for Chinese (\textit{zh}) in XQuAD shown in Table~\ref{tab:xquad}, we observe that the \textit{Act Prob 90p} method outperforms LAPE. This difference in performance can be attributed to the fact that the neurons identified by LAPE and \textit{Act Prob 90p} are largely disjoint, as shown in  Figures~\ref{fig17:llama3_act_layerwise_lang_neurons_dist} and \ref{fig18:mistral_act_layerwise_lang_neurons_dist}.
	
\begin{figure}[ht!]
        \centering
        \includegraphics[width=0.8\linewidth]{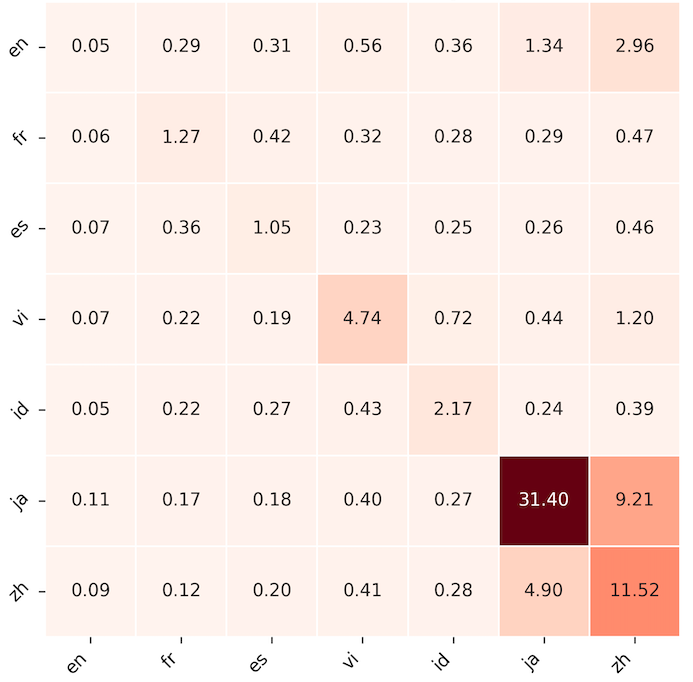} 
        \caption{\footnotesize Perplexity Change (PPXC): Measures the effect of interventions on target language perplexity, defined as $\text{PPXC}(i,j) = \text{PPX}(j \,|\, \text{Intervention by 0 at } i) - \text{PPX}(j)$. Lower $\text{PPXC}(i,j)$ values indicate minimal interference, while higher values signify a significant impact on the model's understanding of language $j$ (on 1 Million tokens).}
        \label{fig:ppx}
    \end{figure}

	\paragraph{Deactivation of Zero Does not Degrade Performance Significantly:}
    Prior studies \cite{12kojima-etal-2024-multilingual, tang-etal-2024-language} commonly deactivate neurons by setting their activations to zero. However, we argue that zero is not necessarily a true indicator of deactivation. While replacing activations with far lower percentiles (such as the 10th percentile) leads to a clear drop in performance (Table~\ref{tab:xquad}), setting activations to zero does not show a similar degradation. Figure~\ref{fig:ppx} illustrates the perplexity change ($\text{PPXC}(i,j)$), defined as the difference in perplexity for language $j$ when language neurons for language $i$ are deactivated versus when they remain active, thereby quantifying the impact of targeted neuron deactivation on language understanding and their role in cross-lingual performance. As illustrated in Figure~\ref{fig:ppx}, deactivation at zero significantly increases perplexity (thus degrading generation quality); however, this degradation in perplexity does not directly translate to a decline in task performance. This suggests that setting activations to zero may not be an effective choice for deactivation. Detailed experimentation results can be found in Appendix \ref{app:C}.

	\subsection{Impact of Neuron Fine-Tuning}
	We fine-tuned the identified language-specific neurons using LoRA but observed no improvement in performance (Table~\ref{tab:ft}). When applying test-time interventions to the fine-tuned models, the results remained consistent with the zero-shot transfer (Table~\ref{tab:xnli}), reinforcing that fine-tuning language neurons does not enhance task performance. We also fine-tuned randomly selected neurons in the MLP layers. The results were similar to both language neuron fine-tuning and the original model (Table ~\ref{tab:rft}), indicating that LoRA applied to attention layers is already effective for task-specific tuning.
	
	\begin{table}[ht!]
		\centering
		 \renewcommand{\arraystretch}{1.15} 
		\setlength{\tabcolsep}{4.5pt} 
		\footnotesize
		\begin{tabular}{|c|c|c|c|c|c|c|}
			\hline
			\textbf{FTL} & \textbf{EL} & \textbf{No Int} & \textbf{Int} $\mathbf{\mu}$ & \textbf{Int P90} & \textbf{Int 0} & \textbf{Int P10} \\ 
			\hline
			\multicolumn{7}{|c|}{\textit{Llama 3.1 with LAPE}} \\ 
			\hline
			en & vi & \textbf{80.2} & 79.6 & 78.5 & 79.2 & 78.0 \\ 
			vi & vi & \textbf{80.1} & 79.5 & 78.6  & 79.2 & 78.0 \\  
			en+vi & vi & \textbf{80.1} & 79.4 & 78.5 & 79.1 & 78.0 \\
			\hline
			en & hi & \textbf{74.9} & 74.6 & 74.6 & 74.1  & 74.6 \\ 
			hi & hi & \textbf{74.9} & 74.6 & 74.5 &  74.3 & 74.6 \\  
			en+hi & hi & \textbf{74.9} & 74.5 & 74.5 &  74.3 & 74.7 \\
			\hline
			en & ur & 69.8 & \textbf{70.4} & 69.6 &  70.2 & 69.0 \\ 
			ur & ur & \textbf{69.8} & 70.5 & 69.5 & 70.4 & 69.1 \\  
			en+ur & ur & 70.0 & \textbf{70.6} & 69.5 & 70.3 & 68.9 \\
			\hline
		\end{tabular}
		\caption{\footnotesize Fine-tuning results for language-specific neurons on XNLI. The results follow the same format as Table~\ref{tab:xnli}, comparing zero-shot performance with test-time interventions across different fine-tuning language neuron (FTL) as per Section \ref{sec:lora}. A complete version is provided in Table~\ref{tab:ft1}.}
		\label{tab:ft}
	\end{table}
	
	\section{Related Works}
    Other from \citet{tang-etal-2024-language}, \citet{zhu-etal-2024-landermt} also introduce LANDeRMT that routes language-aware neurons to mitigate catastrophic forgetting, improving translation quality. Similarly, \citet{xie-etal-2021-importance} propose a neuron allocation strategy to balance general and language-specific knowledge, thereby enhancing translation without increasing complexity. \citet{lai-etal-2024-style} present Neuron-TST, which enhances text style transfer by identifying and deactivating source-style neurons to guide target-style generation. \citet{kojima-etal-2024-multilingual} analyze language-specific neurons in decoder PLMs, showing that manipulating a small subset can control output language. \citet{huo-etal-2024-mmneuron} study domain-specific neurons in Multimodal LLMs, showing a 10\% accuracy gain in domain-specific tasks through neuron manipulation, akin to language-specific neuron use. \citet{durrani-etal-2020-analyzing} analyze encoder models, and find small neuron subsets capture linguistic tasks, with lower-level tasks requiring fewer neurons. No prior work has examined the effect of language-specific neurons on cross-lingual downstream tasks, which we attempt in this work. 

	\section{Conclusion}
	In this work, we investigate if language-specific neurons in multilingual LLMs could be manipulated to improve cross-lingual task performance. Our results show that test-time interventions and fine-tuning of language-specific neurons do not yield meaningful improvements. Altering these activations often disrupt task-relevant information likely due to the polysemantic nature of LLM neurons. We found the same behaviour across different methods for identifying language neurons, such as LAPE and activation probability. Additionally, setting activations to zero did not significantly degrade performance, suggesting that zero is not a true indicator of deactivation. These findings indicate that language-specific neurons do not function independently but interact with broader model components, and need further investigation as tools of cross-lingual transfer. 
    \looseness=-1

	\section*{Limitations}
	This work focuses on language-specific neurons in the MLP layers of multilingual LLMs, excluding attention mechanisms, which may also play a significant role. The experiments use a limited number of languages and datasets, limiting the generalizability of the findings. The interventions rely on statistical activations computed from Wikipedia text, which might not fully capture task-specific behavior. Additionally, the study does not explore alternate methods of fine-tuning techniques that might yield better results. Factors beyond language-specificity in neurons such as training data quality and architectural details of models should also be closely examined for effective cross-lingual transfer.

    \section*{Acknowledgments}
    We are grateful to the anonymous reviewers
    for their insightful feedback. The last author gratefully acknowledges the generous support provided by the joint AI/ML initiative of Amazon and the Indian Institute of Technology Bombay.
	
	\bibliography{main}
	
	\appendix

    \section{Language Neuron Identification}\label{app:A}
	
	\subsection{Dataset Collection and Preprocessing}
	The dataset is constructed from publicly available Wikipedia dumps, specifically from the \citet{20wikidump} dataset. For each language, the following preprocessing steps are applied:
	\begin{itemize}
		\item The dataset is randomly shuffled to ensure diverse text coverage.
		\item Only the first 100 million tokens per language are retained for computational efficiency.
		\item Each sequence is truncated to a maximum context length of $T_{\max} = 512$ tokens.
	\end{itemize}
	
\subsection{Activation Computation}

To identify language-specific neurons, we compute activation statistics from the Wikipedia dataset for each language. This involves calculating both the mean activation and the 90th percentile activation (\textit{P90}) for every neuron in the model. These statistics provide insights into how neurons behave across different languages and form the basis for selecting language-specific neurons.

\subsubsection{Mean Activation Computation}

For a given language $l$, let the activation of neuron $(i, j)$ at token position $t$ in sequence $s$ be denoted as $h_{i,j}^{(s,t)} \in \mathbb{R}$. The mean activation of a neuron across all token positions for a single sequence is computed as:

\begin{equation*}
\bar{h}_{i,j}^{(s)} = \frac{1}{T} \sum_{t=1}^{T} h_{i,j}^{(s,t)},
\end{equation*}

where $T$ represents the sequence length (maximum of 512 tokens). To obtain the overall mean activation for a language $l$, we aggregate across all sequences $S_l$ in the Wikipedia dataset:

\begin{equation*}
\mu_{i,j}^{l} = \frac{1}{|S_l|} \sum_{s \in S_l} \bar{h}_{i,j}^{(s)}.
\end{equation*}

This provides a language-specific average activation for each neuron, which helps in identifying neurons that consistently activate for a particular language.

\subsubsection{90th Percentile Activation Computation}

In addition to mean activation, we compute the 90th percentile activation (\textit{P90}) to capture the upper range of neuron activity. The 90th percentile is useful in determining neurons that are highly responsive in a given language. The \textit{P90} activation for neuron $(i, j)$ in language $l$ is computed as:

\begin{equation*}
P_{90}(h_{i,j}^{l}) = \inf \left\{ x \,|\, F_{h_{i,j}^{(s)}}(x) \geq 0.90, s \in S_l \right\},
\end{equation*}

where $F_{h_{i,j}^{(s)}}(x)$ is the cumulative distribution function (CDF) of the activations of neuron $(i, j)$ for all sequences in language $l$. In practice, this is computed by sorting activation values for all sequences and selecting the value at the 90th percentile position.

\subsection{LAPE and Act Prob 90p Details}
In our experiments, we focus on two specific language sets: \textit{Set1} and \textit{Set6}. Although we have explored different combinations of language sets during our analysis, for clarity and brevity we present results corresponding only to \textit{Set1} and \textit{Set6}.

\textbf{Set1: Core Languages.} This set consists of languages that were part of the original LAPE analysis \cite{tang-etal-2024-language}. It includes: \{\text{en}, \text{fr}, \text{es}, \text{vi}, \text{id}, \text{ja}, \text{zh}\}. This selection covers a broad range of linguistic families, including \textit{Indo-European (en, fr, es), Austroasiatic (vi), Austronesian (id), Japonic (ja), and Sino-Tibetan (zh)}. These languages are well-represented in large-scale multilingual corpora and serve as strong candidates for evaluating multilingual neuron activations.

\textbf{Set6: Indian Language-Dominant Set.} While \textit{Set1} includes a mix of global languages, \textit{Set6} is specifically designed to focus on \textit{Indian languages}: \{\text{en}, \text{bn}, \text{hi}, \text{ta}, \text{te}, \text{mr}, \text{ur}, \text{kn}, \text{ml}, \text{pa}\}. The motivation behind selecting \textit{Set6} is to investigate how LLMs encode representations for typologically and script-wise diverse Indian languages. The inclusion of \textit{Bengali (bn), Hindi (hi), Tamil (ta), Telugu (te), Marathi (mr), Urdu (ur), Kannada (kn), Malayalam (ml), and Punjabi (pa)} ensures a wide coverage of Indo-Aryan and Dravidian language families. 

The LAPE method is evaluated on both \textit{Set1} and \textit{Set6} to determine how neuron activations vary across these two distinct sets. Since \textit{Set1} was originally introduced in prior work, our experiments on \textit{Set6} extend the understanding of LAPE to Indian languages, which are underrepresented in pretrained LLMs.

For the \textit{Activation Probability 90p} (Act Prob 90p) method, we select \{\text{en}, \text{vi}, \text{hi}, \text{ur}, \text{zh}\}. This selection was based on language diversity, cross-lingual representation, and performance disparities in downstream tasks. Since Act Prob 90p is a set-independent method, we focus on selecting languages that exhibit poor performance in task-specific evaluations. Specifically, for the XNLI task, the lowest-performing languages were \emph{Hindi (hi), Urdu (ur), and Vietnamese (vi)}, leading to their inclusion. Similarly, for the XQuAD task, the weakest-performing languages were \emph{Vietnamese (vi), Hindi (hi), and Chinese (zh)}, which motivated their selection.

\section{Task, Models and Experiment Details}\label{app:B}
    
In this section, we provide detailed descriptions of the two evaluation tasks used in our experiments, namely XNLI and XQuAD, as well as the two large language models (LLMs) used for our study: Llama 3.1 and Mistral Nemo. We also formalize the task setup using mathematical notations.

\subsection{Tasks}

\subsubsection{XNLI}

The XNLI dataset \cite{15conneau2018xnlievaluatingcrosslingualsentence} is a cross-lingual extension of the MultiNLI dataset, designed for evaluating natural language inference (NLI) across multiple languages. Given a premise $p$ and a hypothesis $h$, the task is to determine whether the hypothesis is \textit{entailment}, \textit{contradiction}, or \textit{neutral} with respect to the premise. Formally, given a dataset $\mathcal{D} = \{(p_i, h_i, y_i)\}_{i=1}^{N}$, where: $p_i \in \mathcal{X}$ is the premise, $h_i \in \mathcal{X}$ is the hypothesis, $y_i \in \{0,1,2\}$ represents the label: entailment (0), contradiction (1), or neutral (2). We conduct zero-shot evaluation on target languages (\textit{vi}, \textit{hi}, \textit{ur}), using English (\textit{en}) as the source language. We limit the training dataset to 100,000 samples (25\% of the full dataset) for efficiency.

\subsubsection{XQuAD: Cross-lingual Question Answering}

XQuAD \cite{16Artetxe_2020} is a multilingual question-answering dataset based on the Stanford Question Answering Dataset (SQuAD). The task requires extracting an answer span $a$ from a given context $c$ for a question $q$. Given a dataset $\mathcal{D} = \{(c_i, q_i, a_i)\}_{i=1}^{M}$, where: $c_i \in \mathcal{X}$ is the passage (context), $q_i \in \mathcal{X}$ is the question, $a_i \in \mathcal{X}$ is the ground-truth answer. We evaluate on target languages (\textit{vi}, \textit{hi}, \textit{zh}) and use the full training dataset for fine-tuning.

\subsection{Models}

\subsubsection{Llama 3.1}

Llama 3.1\footnote{\url{https://huggingface.co/meta-llama/Llama-3.1-8B}} is an 8 billion parameter multilingual model from Meta, trained on diverse text corpora across multiple languages \cite{5grattafiori2024llama3herdmodels}. The model consists of stacked transformer layers, each comprising self-attention and feedforward MLP components. Llama 3.1 is optimized for computational efficiency and supports a wide range of languages, making it a strong candidate for evaluating multilingual transfer performance.

\subsubsection{Mistral Nemo}

Mistral Nemo\footnote{\url{https://huggingface.co/mistralai/Mistral-Nemo-Base-2407}} is a 12 billion parameter transformer based model designed for multilingual tasks, with a particular emphasis on high-performance fine-tuning capabilities \cite{6mistral-nemo}. Similar to Llama 3.1, it consists of transformer layers with self-attention and MLP modules. 

\subsection{Implementation Details for LoRA Fine-Tuning}

In this section, we provide an overview of the implementation details for our fine-tuning experiments on the XNLI and XQuAD tasks using LoRA.

For model configuration, our experiments were conducted using the Meta-Llama-3.1-8B and Mistral-Nemo-Base-2407 models. Both models were loaded in 4-bit precision to optimize efficiency. Specifically, we employed the \texttt{nf4} quantization type, used \texttt{bfloat16} as the compute data type, and enabled double quantization.

Regarding task-specific dataset preparation, for the XNLI dataset—which involves natural language inference by predicting entailment, contradiction, or neutral relationships—we used 25\% of the training data and 100\% of the evaluation data. The maximum context length was set to 256 tokens. For the XQuAD dataset, which focuses on question answering by extracting answer spans from a given context, we utilized the full training data (100\%) along with 100\% of the evaluation data, with a maximum context length of 512 tokens.

LoRA was applied to fine-tune specific layers in the attention module of the models for both tasks. The fine-tuning was performed with a LoRA rank \(r = 64\) and a LoRA scaling factor \(\alpha = 128\). The learning rate was set to \(1 \times 10^{-6}\) for XNLI and \(5 \times 10^{-5}\) for XQuAD, with a weight decay of 0.1 and gradient clipping at a threshold of 10.0. The AdamW optimizer was used with parameters \(\beta_1 = 0.95\) and \(\beta_2 = 0.999\).

For the training configuration, we trained the model for 2 epochs on XNLI using a batch size of 8, and for 10 epochs on XQuAD using a batch size of 4. A linear warm-up was employed for 1\% of the total steps, followed by a linear decay of the learning rate. Mixed precision training was enabled with \texttt{bfloat16} to improve memory efficiency.

\section{Results of Language Neuron Analysis}\label{app:C}

Figure~\ref{fig1:llama3_set1_lang_neuron_dist} to \ref{fig6:mistral_act_lang_neuron_dist} presents the number of neurons assigned per language, comparing the LAPE and Activation Probability 90p methods. The overlap of language-specific neurons across different languages is illustrated in Figure~\ref{fig7:llama3_set1_lang_neurons_overlap} to \ref{fig12:mistral_act_lang_neurons_overlap},  highlighting the extent of shared neurons between languages. The layer-wise distribution of these neurons is shown in Figure~\ref{fig13:llama3_set1_layerwise_lang_neurons_dist} to \ref{fig18:mistral_act_layerwise_lang_neurons_dist}, providing insights into where language-specific representations are most prominent in the model. Finally, the impact of neuron interventions on perplexity is analyzed in Figure~\ref{fig19:llama3_set6_ppx_change} to \ref{fig23:mistral_act_ppx_change}, which displays the perplexity change across different languages when language neurons are manipulated. These figures collectively summarize the key findings from our language neuron analysis.

Table~\ref{tab:xnli1} presents the full XNLI results, extending the analysis from Table~\ref{tab:xnli}, incorporating additional statistical interventions, including percentiles at P75, P90, P95, P5, P10, and P25. Similarly, Table~\ref{tab:xquad1} provides the full XQuAD results, expanding upon Table~\ref{tab:xquad}, detailing both exact match (EM) and F1 scores across various intervention methods. The complete activation statistics for both Llama 3.1 and Mistral Nemo are listed in Table~\ref{tab:act}, offering a breakdown of mean activations and quantiles, capturing the variations in neuron activations across languages. Finally, Table~\ref{tab:ft1} details the full language neuron fine-tuning results, extending Table~\ref{tab:ft}, comparing zero-shot performance with fine-tuning on different language neuron setups, and evaluating test-time interventions across multiple configurations.

\iftrue
\begin{figure}[ht!]
    \centering
    \includegraphics[width=0.45\textwidth]{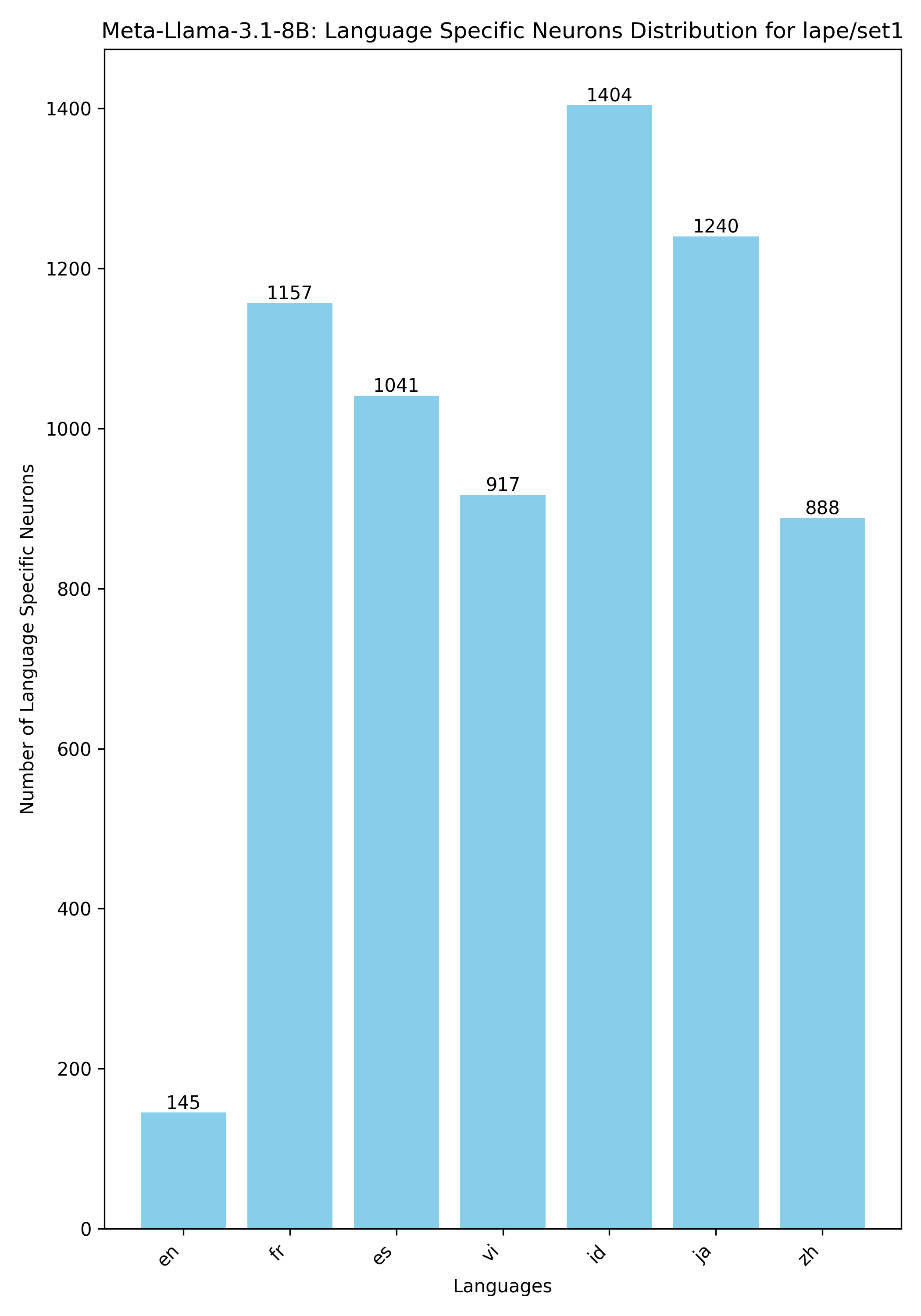}
    \caption{\footnotesize Llama 3.1: Number of language neurons assigned per language for LAPE in a set of languages \{\textit{en},\textit{es},\textit{fr},\textit{vi},\textit{id},\textit{zh},\textit{ja}\}.}
    \label{fig1:llama3_set1_lang_neuron_dist}
\end{figure}
\begin{figure}[ht!]
    \centering
    \includegraphics[width=0.45\textwidth]{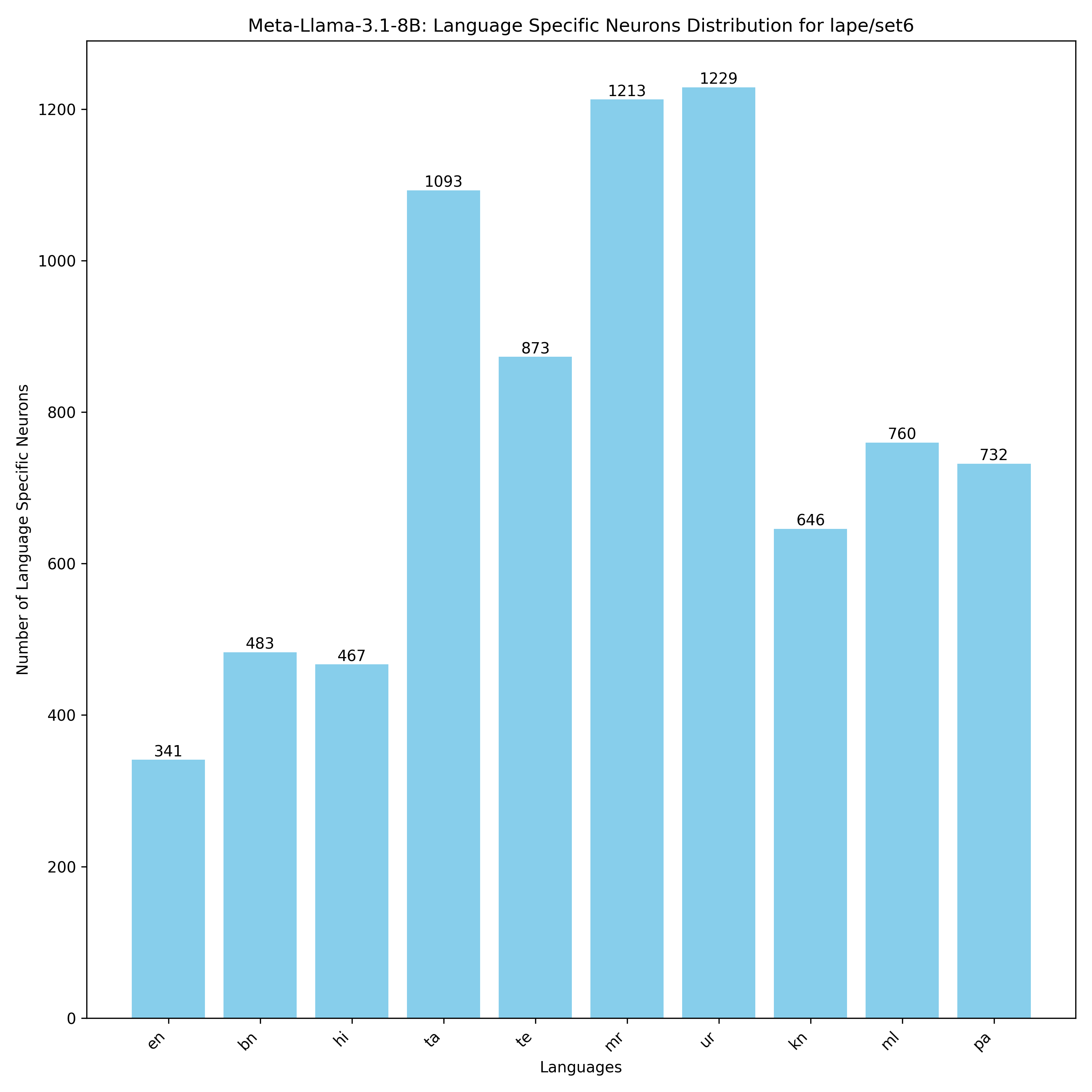}
    \caption{\footnotesize Llama 3.1: Number of language neurons assigned per language for LAPE in a set of languages \{\textit{en},\textit{bn},\textit{hi},\textit{ur},\textit{mr},\textit{pa},\textit{ta}, \textit{te}, \textit{ml}, \textit{kn}\}.}
    \label{fig2:llama3_set6_lang_neuron_dist}
\end{figure}
\begin{figure}[ht!]
    \centering
    \includegraphics[width=0.45\textwidth]{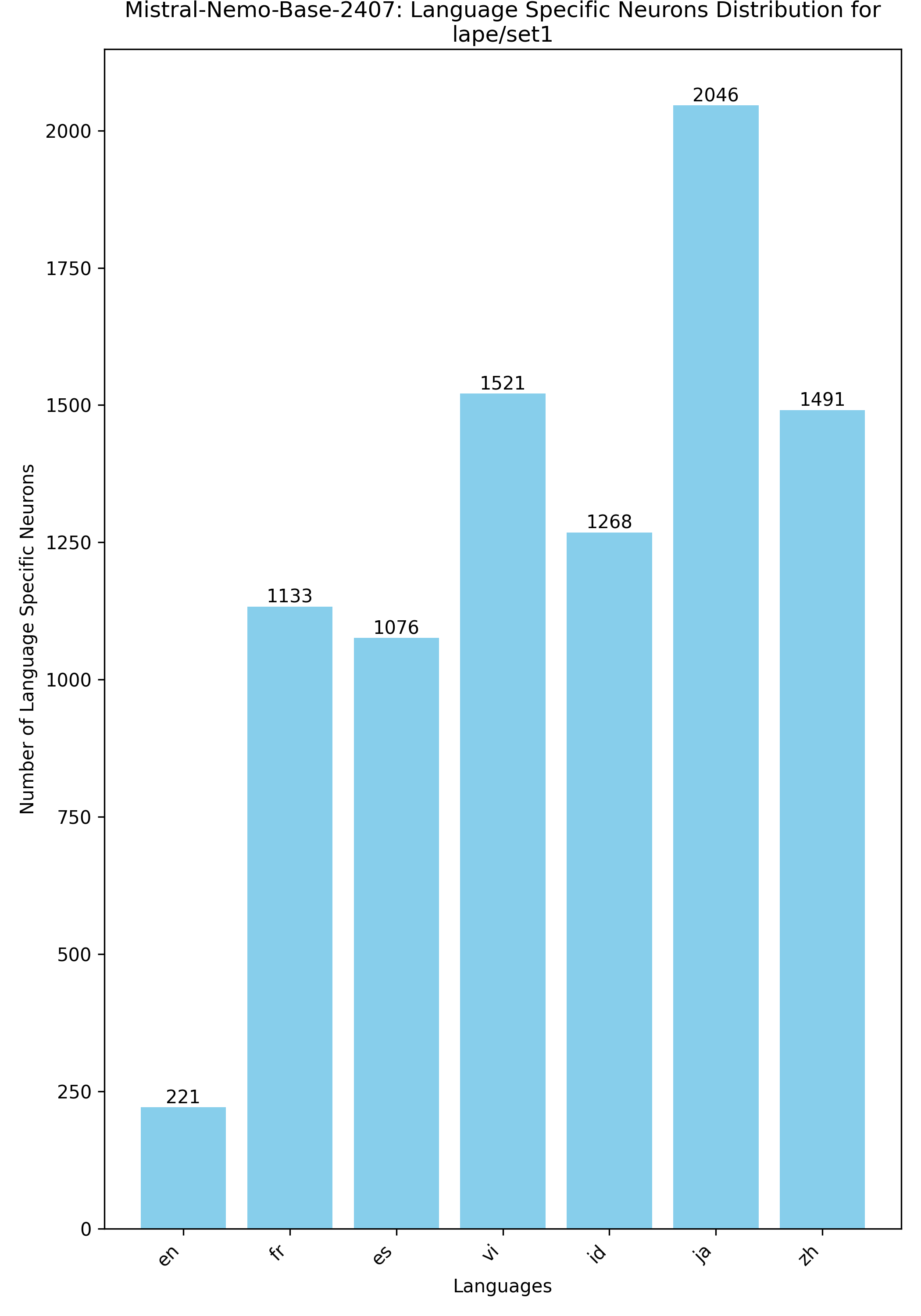}
    \caption{\footnotesize Mistral Nemo: Number of language neurons assigned per language for LAPE in a set of languages \{\textit{en},\textit{es},\textit{fr},\textit{vi},\textit{id},\textit{zh},\textit{ja}\}.}
    \label{fig3:mistral_set1_lang_neuron_dist}
\end{figure}
\begin{figure}[ht!]
    \centering
    \includegraphics[width=0.45\textwidth]{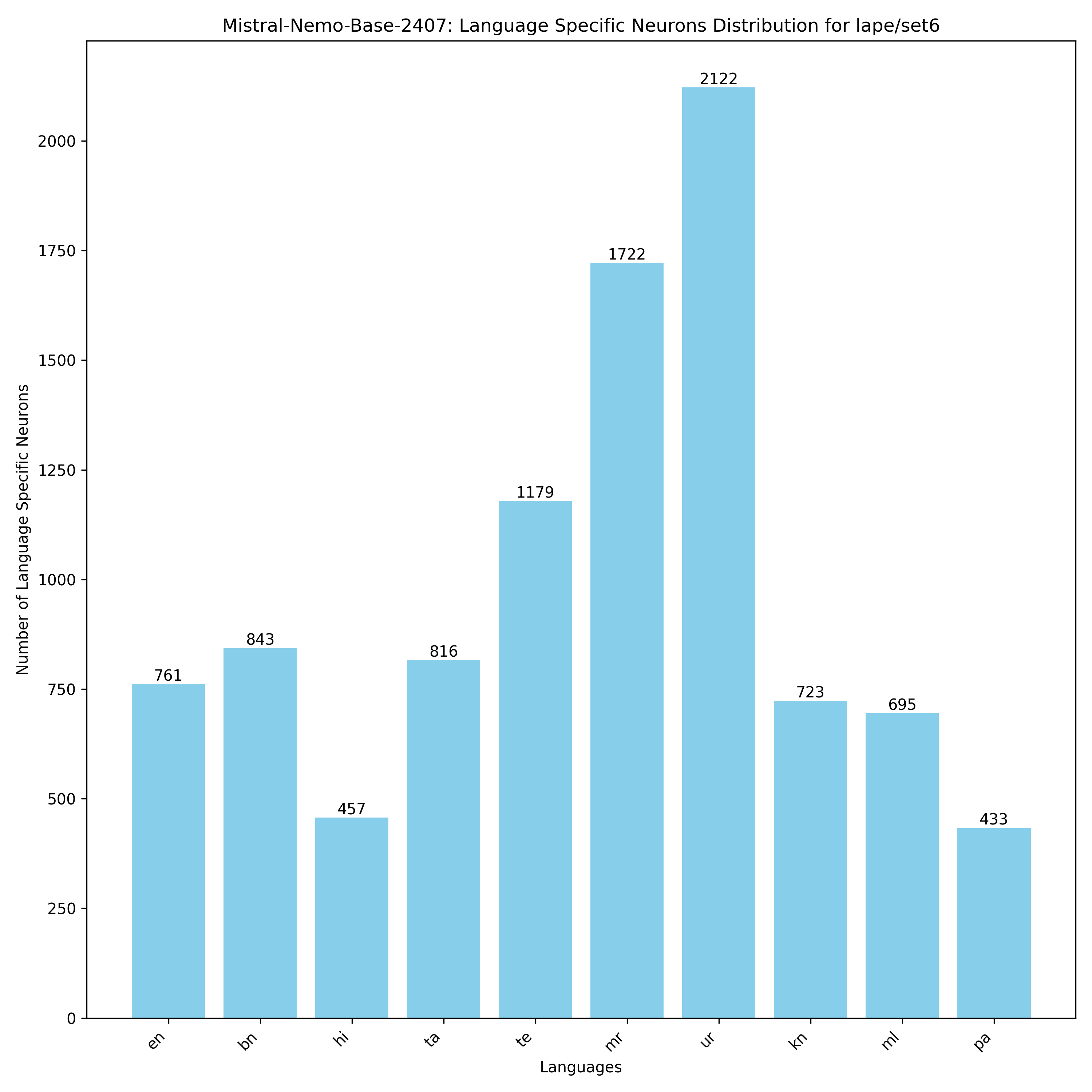}
    \caption{\footnotesize Mistral Nemo: Number of language neurons assigned per language for LAPE in a set of languages \{\textit{en},\textit{bn},\textit{hi},\textit{ur},\textit{mr},\textit{pa},\textit{ta}, \textit{te}, \textit{ml}, \textit{kn}\}.}
    \label{fig4:mistral_set6_lang_neuron_dist}
\end{figure}
\begin{figure}[ht!]
    \centering
    \includegraphics[width=0.2\textwidth]{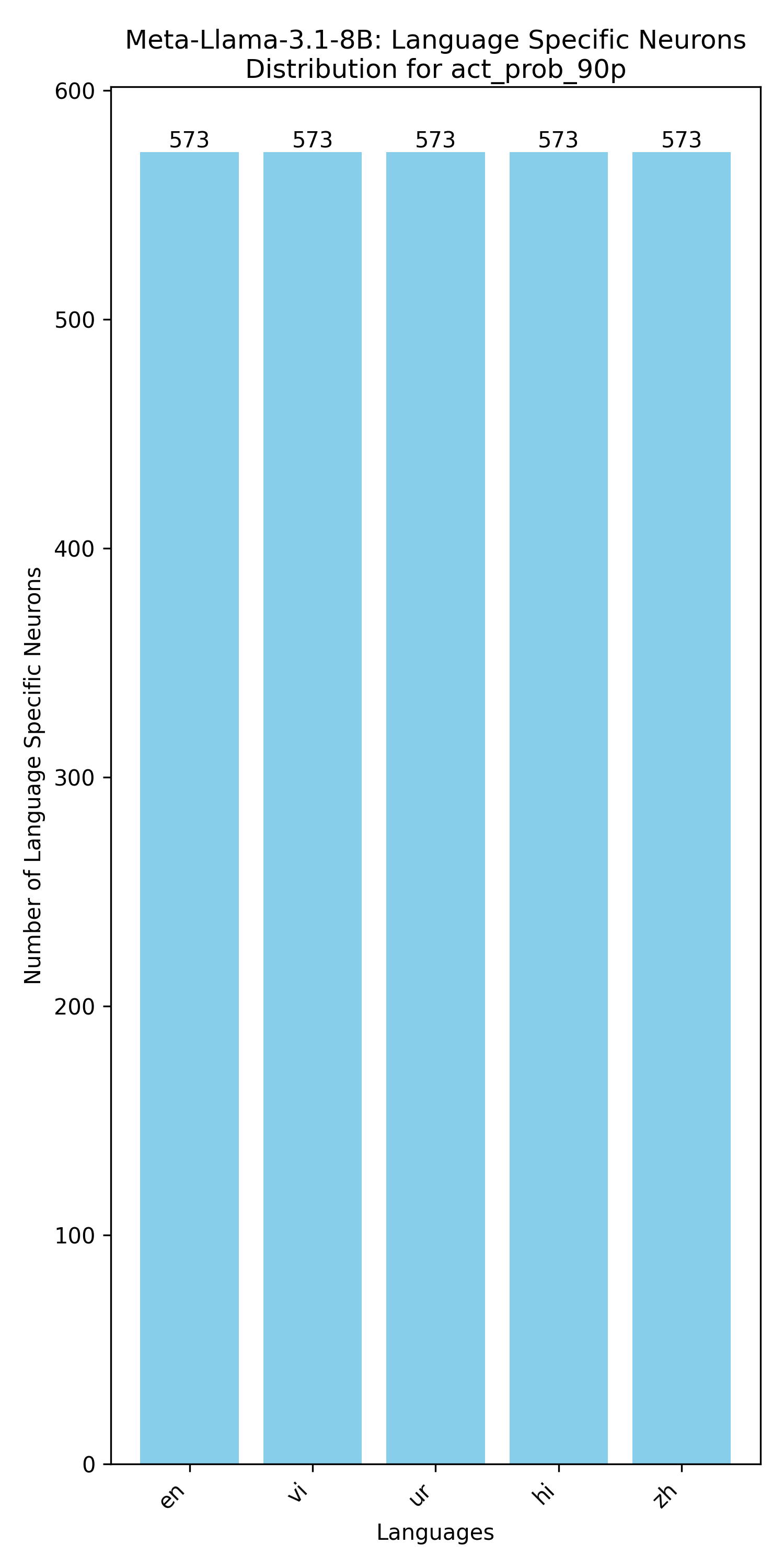}
    \caption{\footnotesize Llama 3.1: Number of language neurons assigned per language for Activation Probability 90p which is same for all the languages.}
    \label{fig5:llama3_act_lang_neuron_dist}
\end{figure}
\begin{figure}[ht!]
    \centering
    \includegraphics[width=0.2\textwidth]{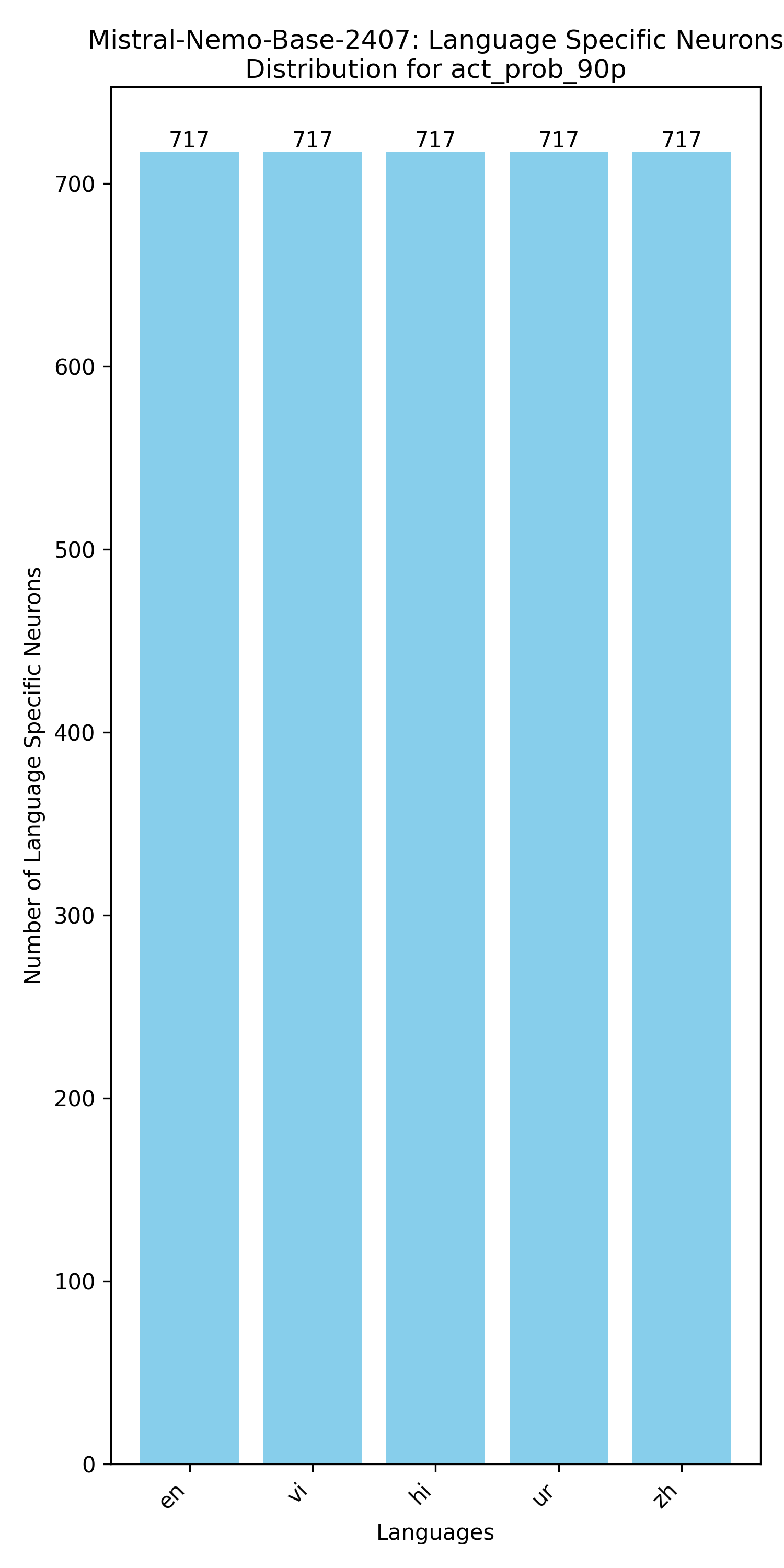}
    \caption{\footnotesize Mistral Nemo: Number of language neurons assigned per language for Activation Probability 90p which is same for all the languages.}
    \label{fig6:mistral_act_lang_neuron_dist}
\end{figure}
\begin{figure}[ht!]
    \centering
    \includegraphics[width=0.45\textwidth]{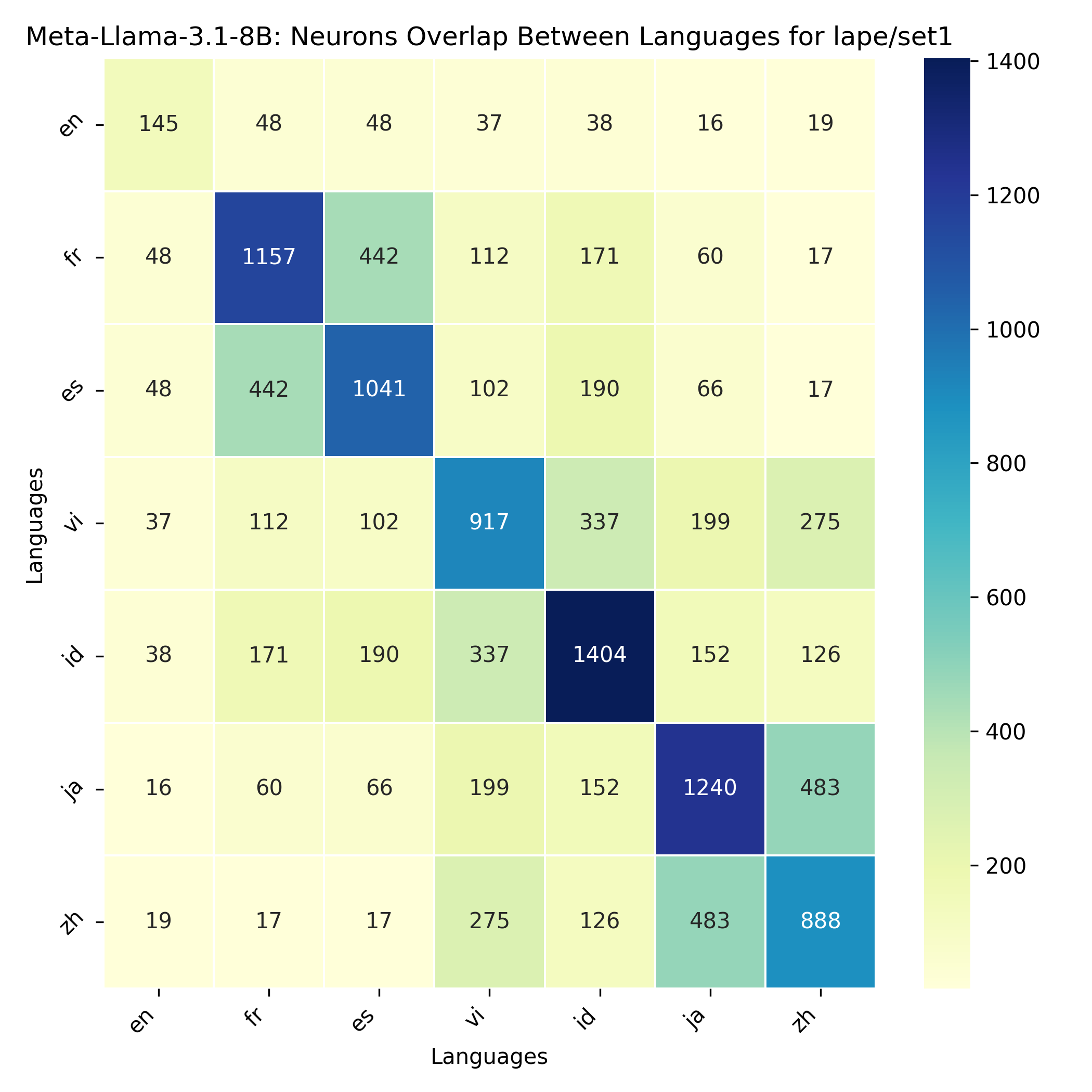}
    \caption{\footnotesize Llama 3.1: Language neuron overlap between languages using LAPE in a set of languages \{\textit{en},\textit{es},\textit{fr},\textit{vi},\textit{id},\textit{zh},\textit{ja}\}.}
    \label{fig7:llama3_set1_lang_neurons_overlap}
\end{figure}
\begin{figure}[ht!]
    \centering
    \includegraphics[width=0.45\textwidth]{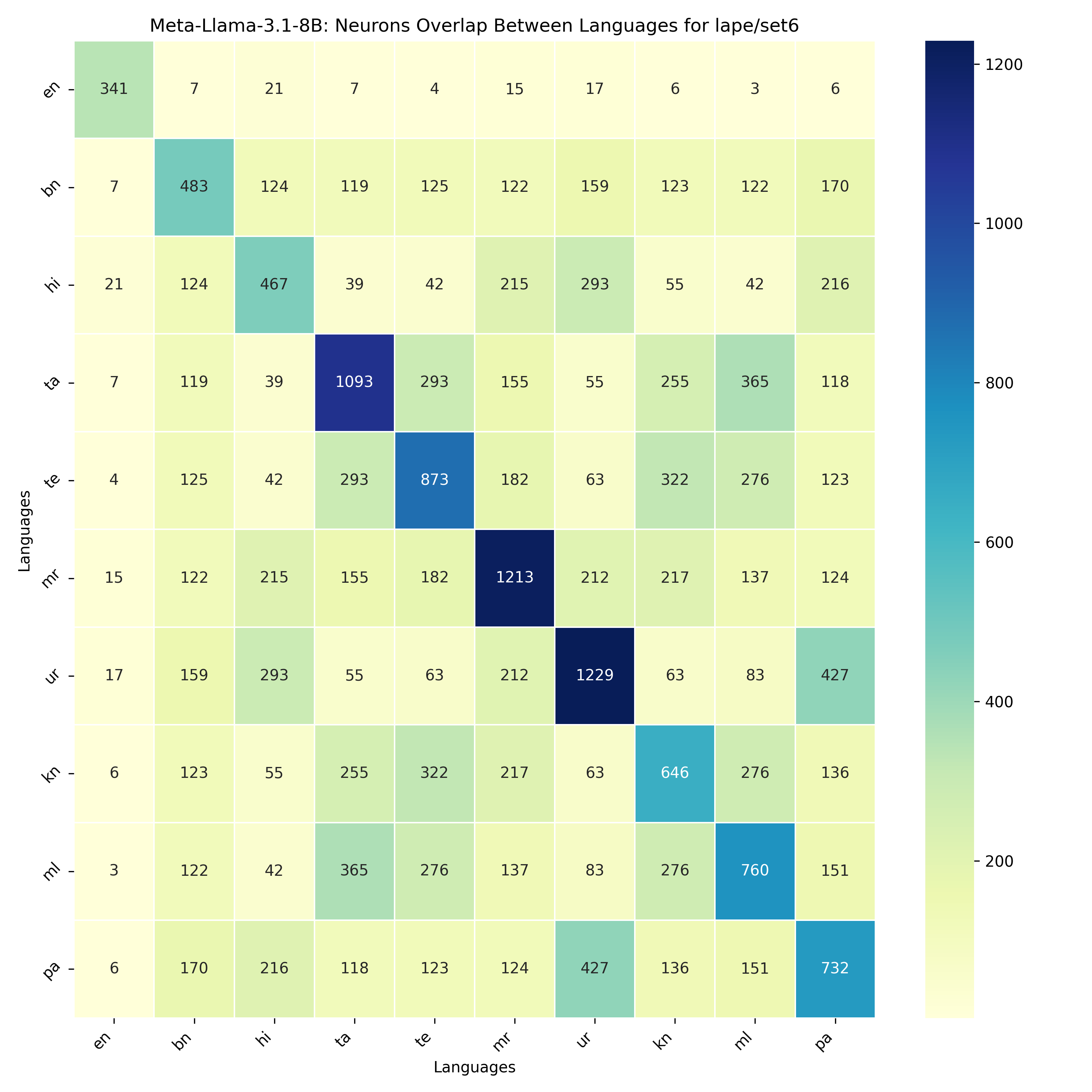}
    \caption{\footnotesize Llama 3.1: Language neuron overlap between languages using LAPE in a set of languages \{\textit{en},\textit{bn},\textit{hi},\textit{ur},\textit{mr},\textit{pa},\textit{ta}, \textit{te}, \textit{ml}, \textit{kn}\}.}
    \label{fig8:llama3_set6_lang_neurons_overlap}
\end{figure}
\begin{figure}[ht!]
    \centering
    \includegraphics[width=0.45\textwidth]{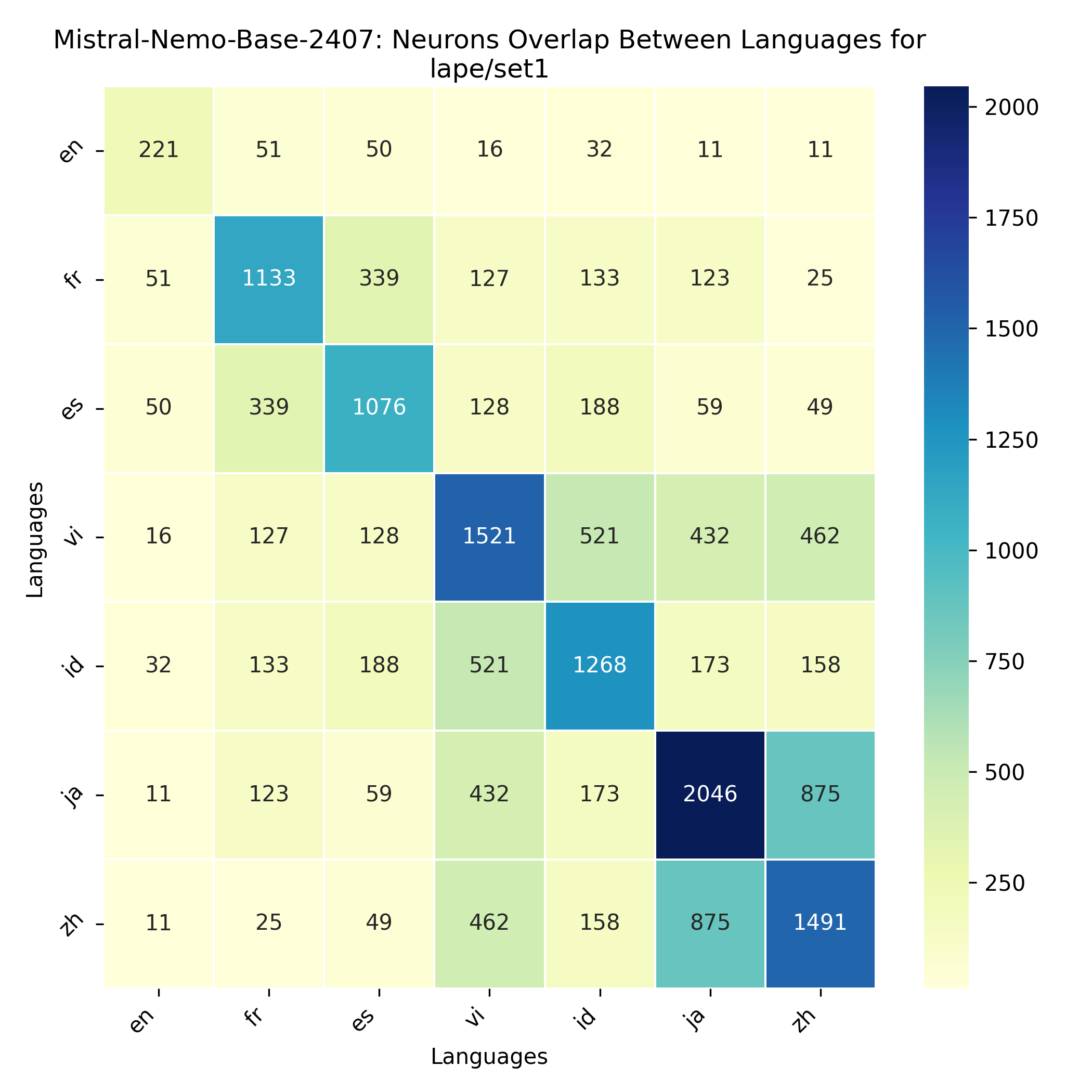}
    \caption{\footnotesize Mistral Nemo: Language neuron overlap between languages using LAPE in a set of languages \{\textit{en},\textit{es},\textit{fr},\textit{vi},\textit{id},\textit{zh},\textit{ja}\}.}
    \label{fig9:mistral_set1_lang_neurons_overlap}
\end{figure}
\begin{figure}[ht!]
    \centering
    \includegraphics[width=0.45\textwidth]{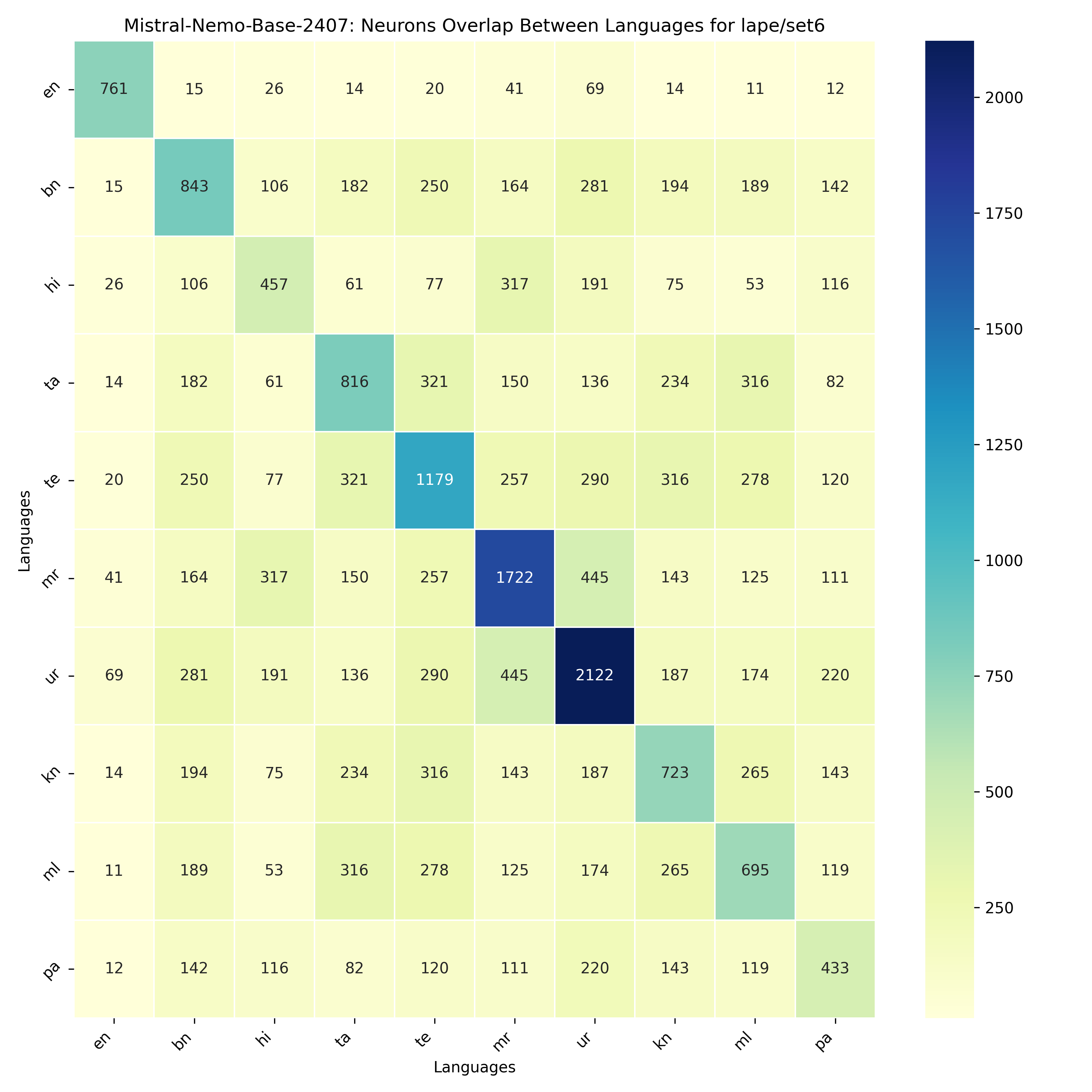}
    \caption{\footnotesize Mistral Nemo: Language neuron overlap between languages using LAPE in a set of languages \{\textit{en},\textit{bn},\textit{hi},\textit{ur},\textit{mr},\textit{pa},\textit{ta}, \textit{te}, \textit{ml}, \textit{kn}\}.}
    \label{fig10:mistral_set6_lang_neurons_overlap}
\end{figure}
\begin{figure}[ht!]
    \centering
    \includegraphics[width=0.45\textwidth]{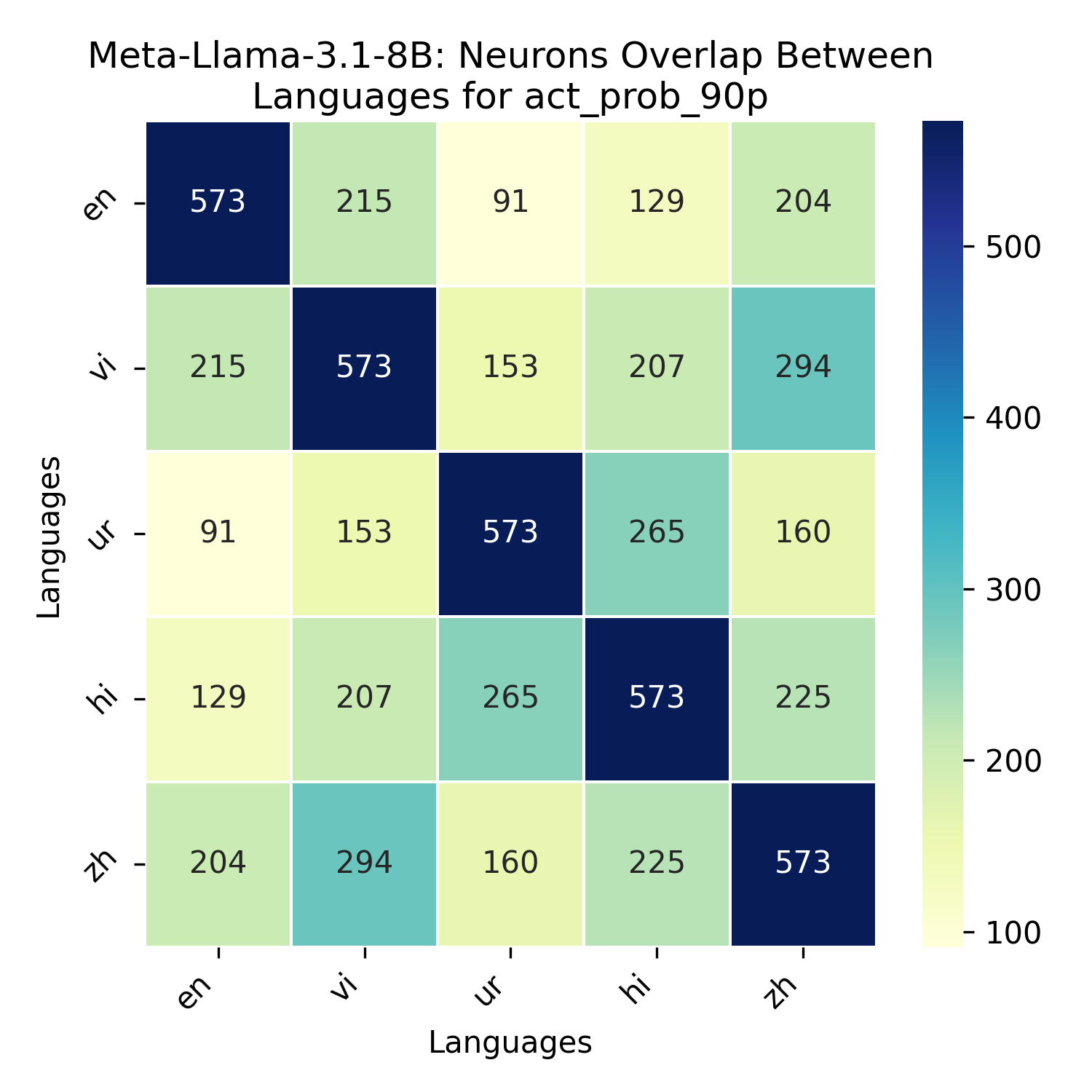}
    \caption{\footnotesize Llama 3.1: Language neuron overlap between languages using Activation Probability 90p.}
    \label{fig11:llama3_act_lang_neurons_overlap}
\end{figure}
\begin{figure}[ht!]
    \centering
    \includegraphics[width=0.45\textwidth]{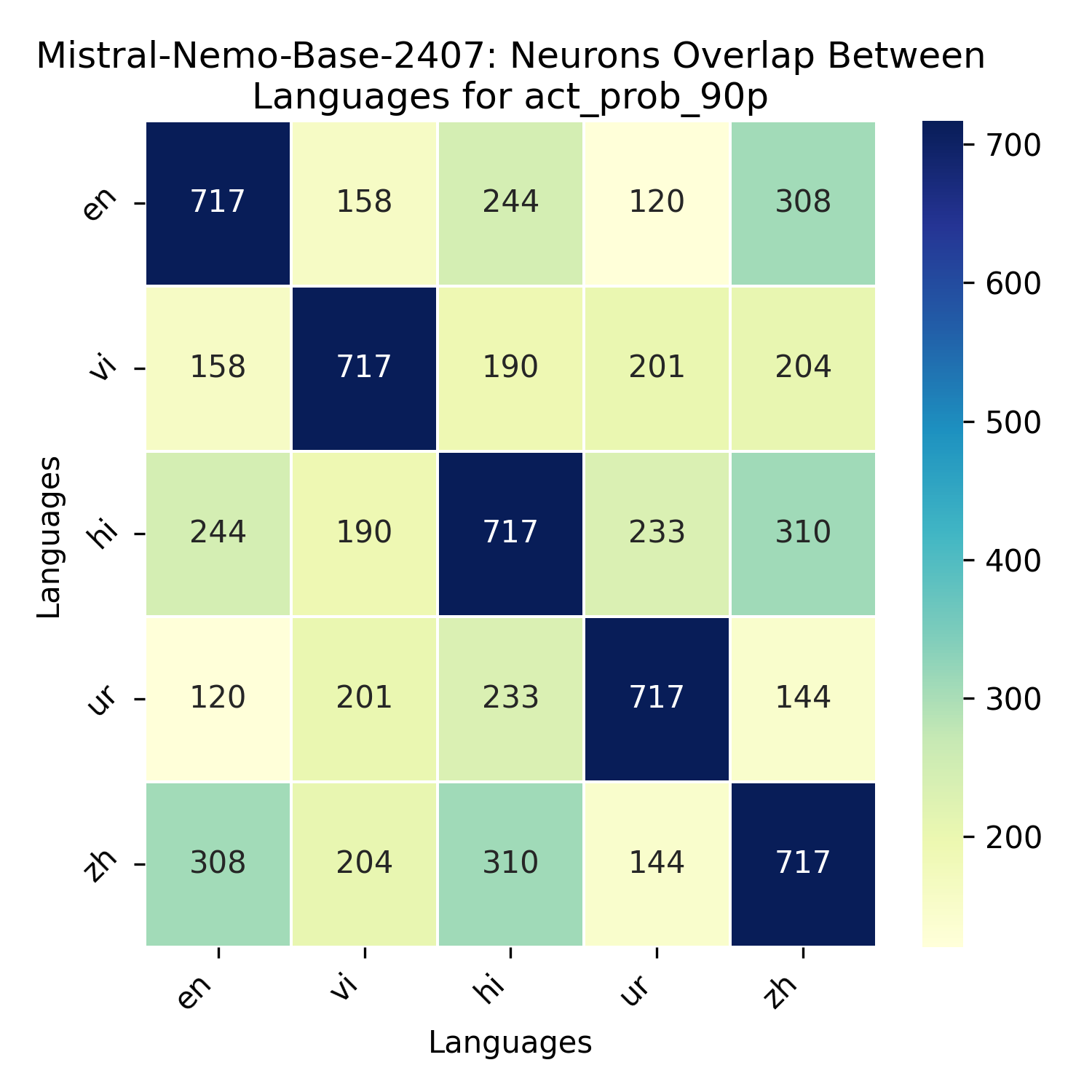}
    \caption{\footnotesize Mistral Nemo: Language neuron overlap between languages using Activation Probability 90p.}
    \label{fig12:mistral_act_lang_neurons_overlap}
\end{figure}
\begin{figure}[ht!]
    \centering
    \includegraphics[width=0.45\textwidth]{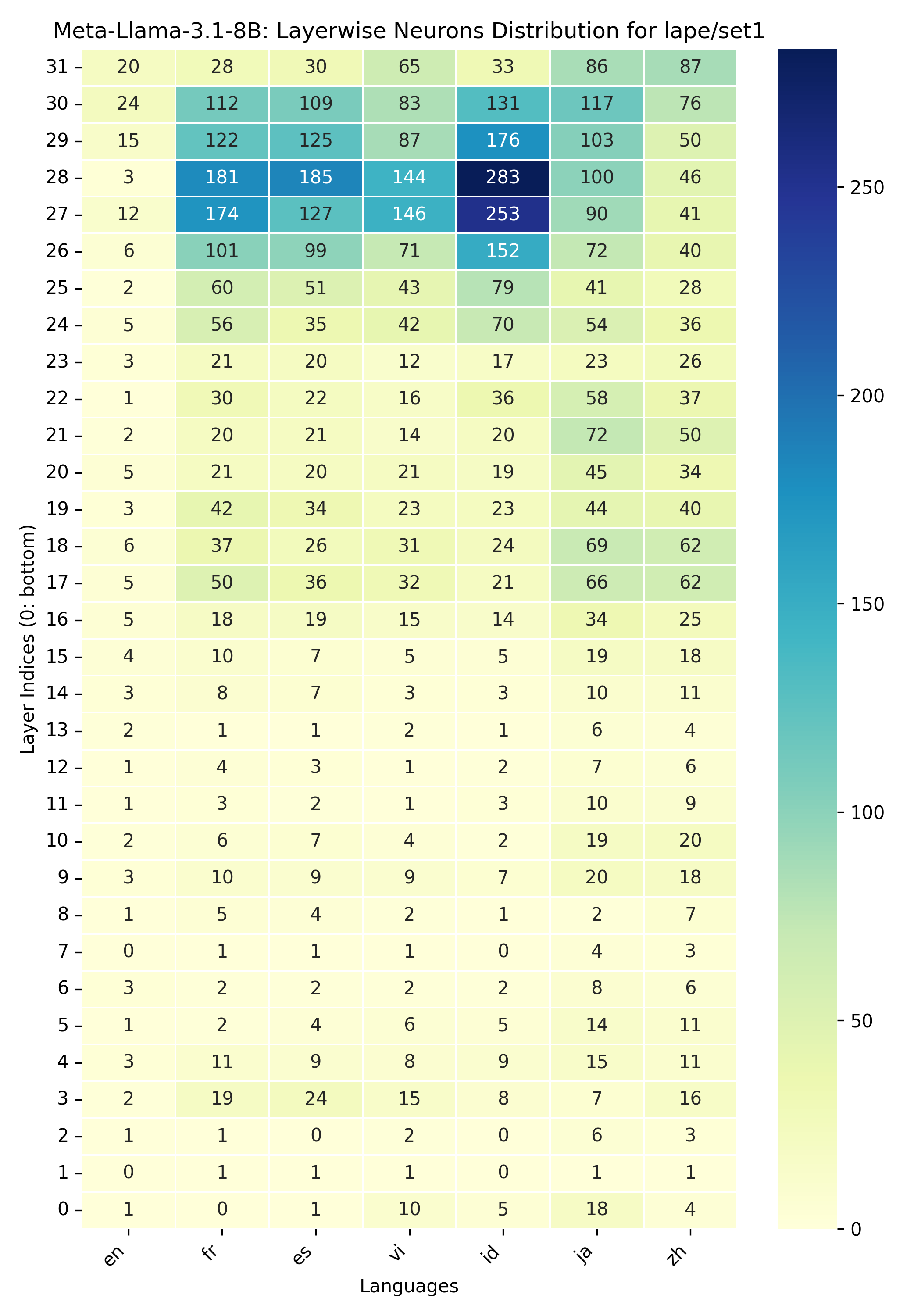}
    \caption{\footnotesize Llama 3.1: Layer-wise distribution of language neurons for LAPE in a set of languages \{\textit{en},\textit{es},\textit{fr},\textit{vi},\textit{id},\textit{zh},\textit{ja}\}.}
    \label{fig13:llama3_set1_layerwise_lang_neurons_dist}
\end{figure}
\begin{figure}[ht!]
    \centering
    \includegraphics[width=0.45\textwidth]{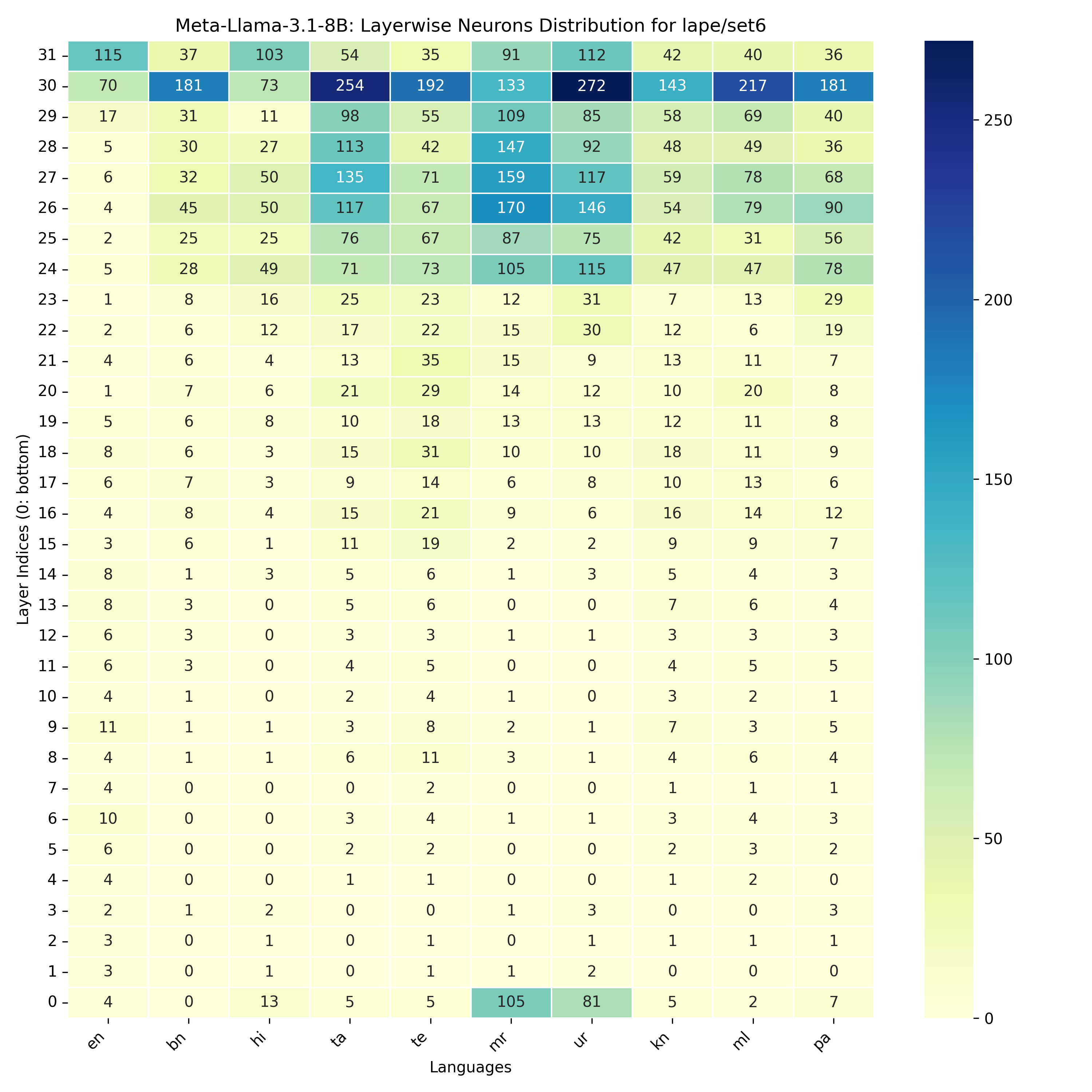}
    \caption{\footnotesize Llama 3.1: Layer-wise distribution of language neurons for LAPE in a set of languages \{\textit{en},\textit{bn},\textit{hi},\textit{ur},\textit{mr},\textit{pa},\textit{ta}, \textit{te}, \textit{ml}, \textit{kn}\}.}
    \label{fig14:llama3_set6_layerwise_lang_neurons_dist}
\end{figure}
\begin{figure}[ht!]
    \centering
    \includegraphics[width=0.45\textwidth]{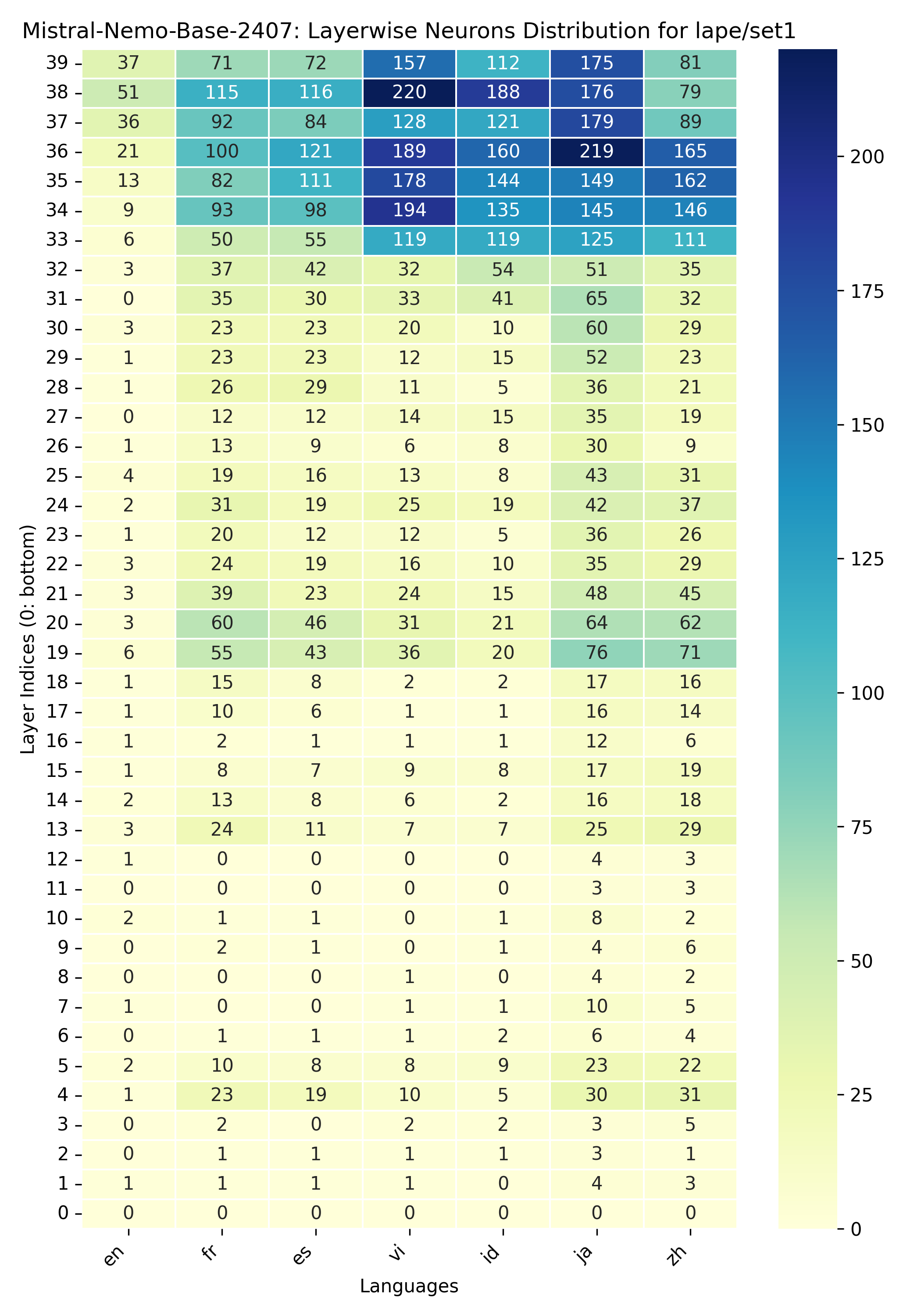}
    \caption{\footnotesize Mistral Nemo: Layer-wise distribution of language neurons for LAPE in a set of languages \{\textit{en},\textit{es},\textit{fr},\textit{vi},\textit{id},\textit{zh},\textit{ja}\}.}
    \label{fig15:mistral_set1_layerwise_lang_neurons_dist}
\end{figure}
\begin{figure}[ht!]
    \centering
    \includegraphics[width=0.45\textwidth]{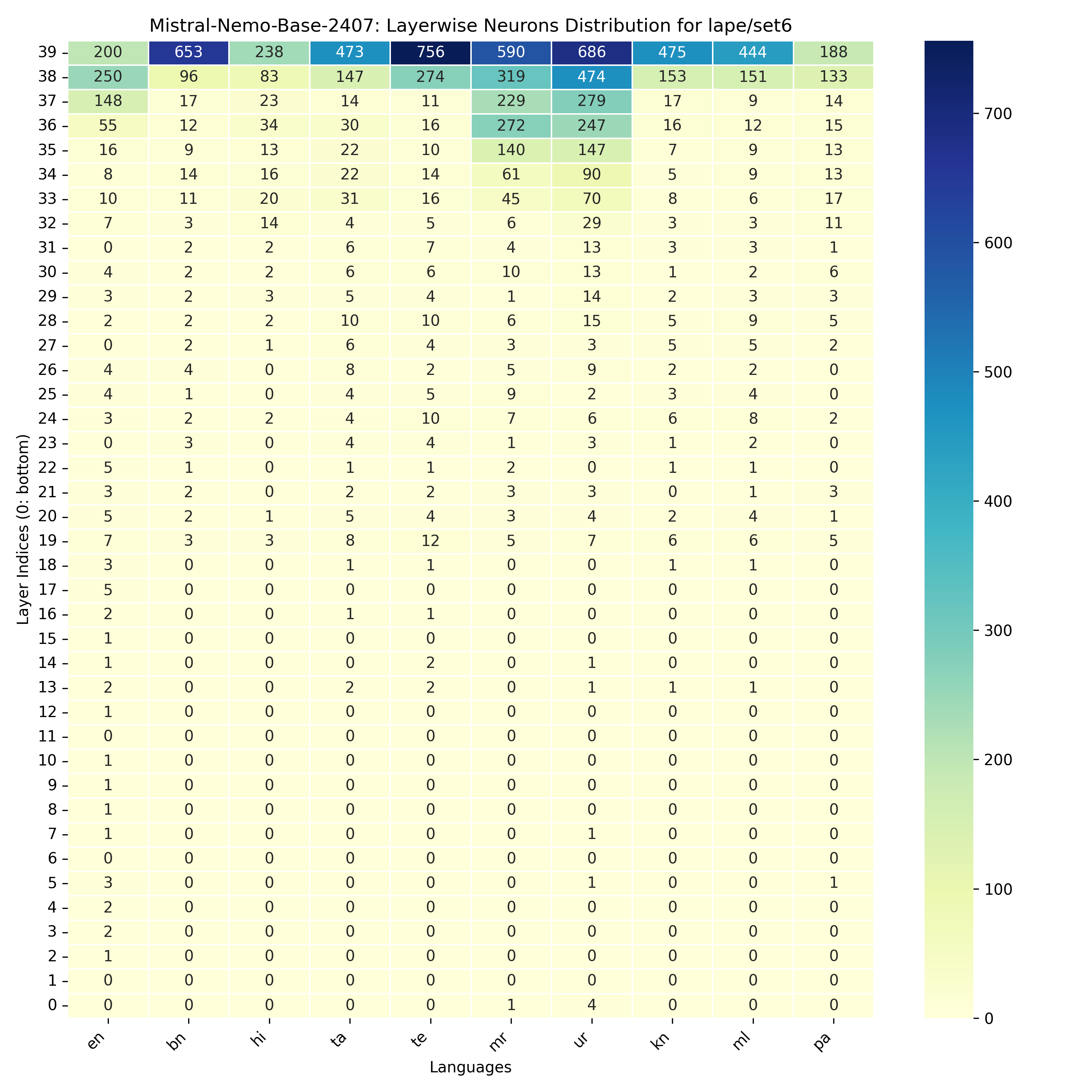}
    \caption{\footnotesize Mistral Nemo: Layer-wise distribution of language neurons for LAPE in a set of languages \{\textit{en},\textit{bn},\textit{hi},\textit{ur},\textit{mr},\textit{pa},\textit{ta}, \textit{te}, \textit{ml}, \textit{kn}\}.}
    \label{fig16:mistral_set6_layerwise_lang_neurons_dist}
\end{figure}
\begin{figure}[ht!]
    \centering
    \includegraphics[width=0.45\textwidth]{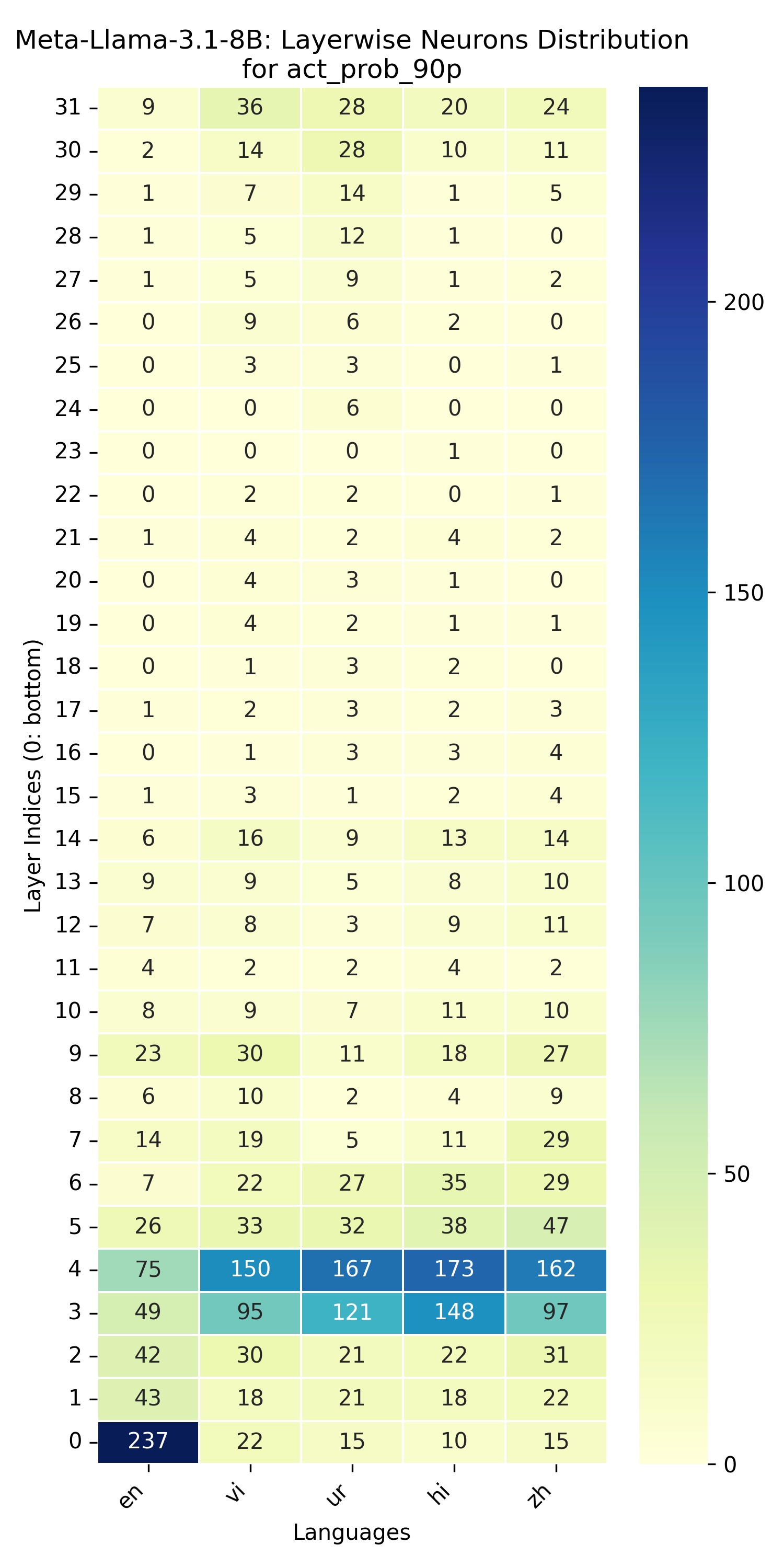}
    \caption{\footnotesize Llama 3.1: Layer-wise distribution of language neurons for Activation Probability 90p.}
    \label{fig17:llama3_act_layerwise_lang_neurons_dist}
\end{figure}
\begin{figure}[ht!]
    \centering
    \includegraphics[width=0.45\textwidth]{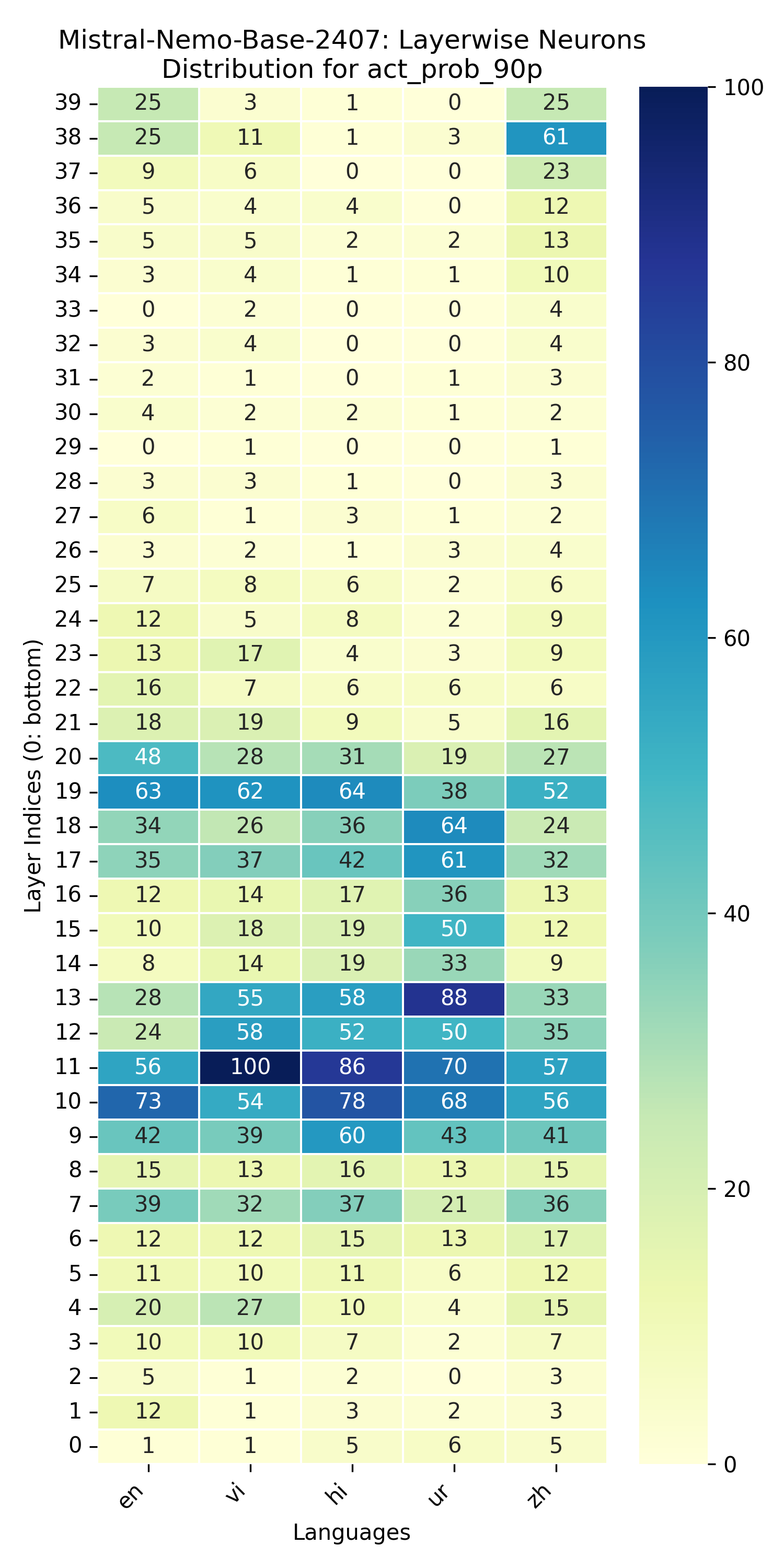}
    \caption{\footnotesize Mistral Nemo: Layer-wise distribution of language neurons for Activation Probability 90p.}
    \label{fig18:mistral_act_layerwise_lang_neurons_dist}
\end{figure}
\begin{figure}[ht!]
	\centering
	\includegraphics[width=0.45\textwidth]{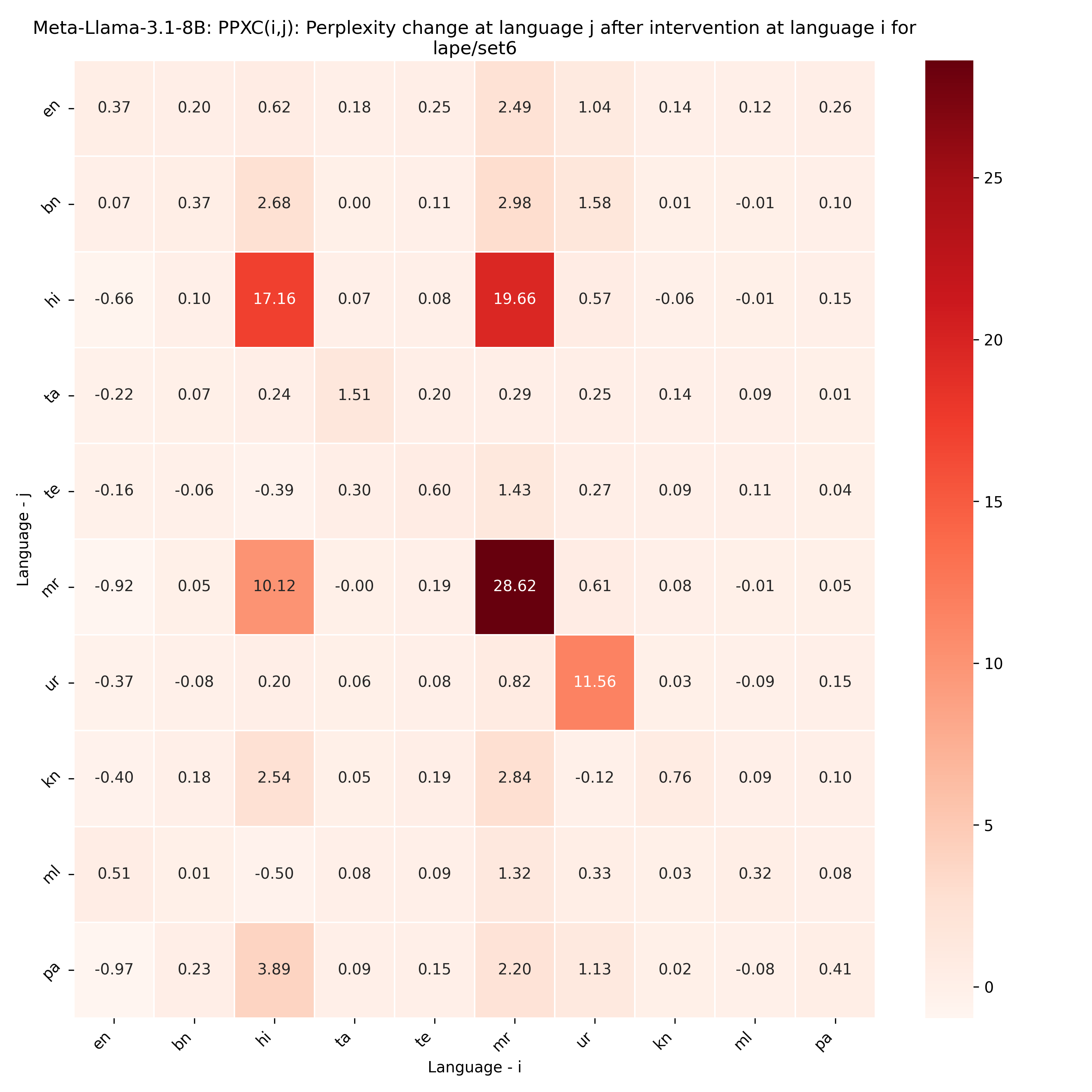}
	\caption{\footnotesize Llama 3.1: Perplexity change for LAPE  in a set of languages \{\textit{en},\textit{bn},\textit{hi},\textit{ur},\textit{mr},\textit{pa},\textit{ta}, \textit{te}, \textit{ml}, \textit{kn}\} (on 0.1 Million tokens).}
	\label{fig19:llama3_set6_ppx_change}
\end{figure}
\begin{figure}[ht!]
	\centering
	\includegraphics[width=0.45\textwidth]{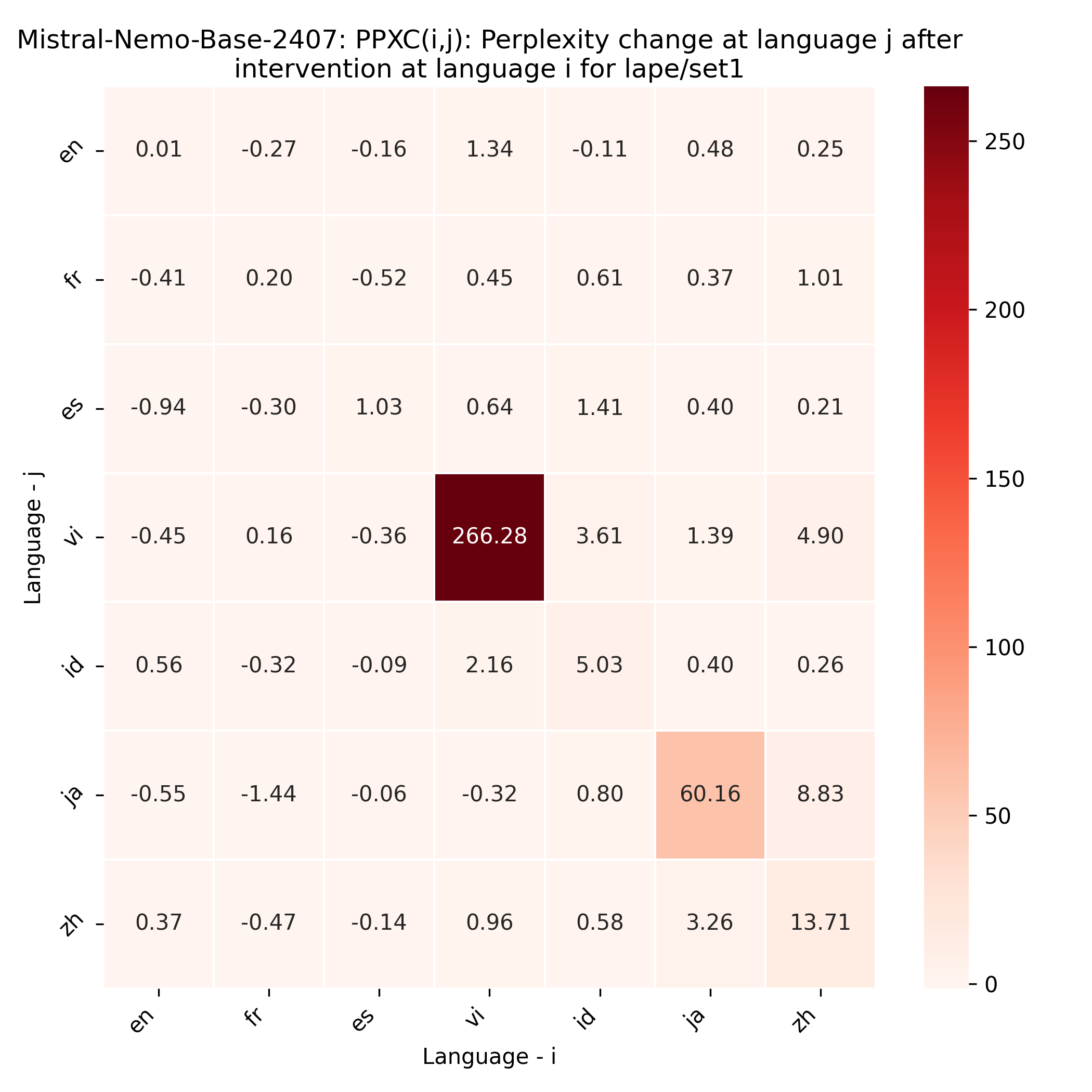}
	\caption{\footnotesize Mistral Nemo: Perplexity change for LAPE in a set of languages \{\textit{en},\textit{es},\textit{fr},\textit{vi},\textit{id},\textit{zh},\textit{ja}\} (on 0.1 Million tokens).}
	\label{fig20:mistral_set1_ppx_change}
\end{figure}
\begin{figure}[ht!]
	\centering
	\includegraphics[width=0.45\textwidth]{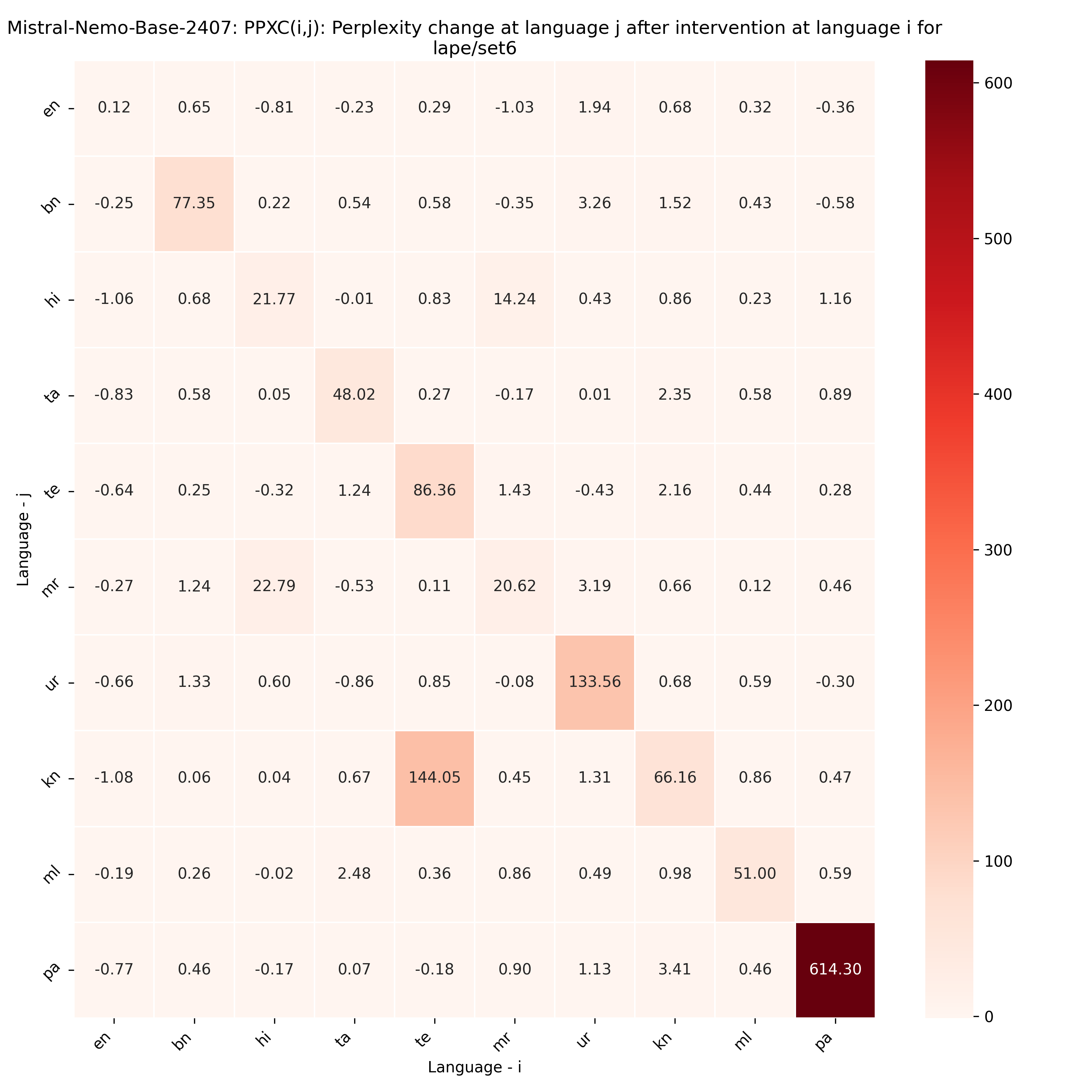}
	\caption{\footnotesize Mistral Nemo: Perplexity change for LAPE  in a set of languages \{\textit{en},\textit{bn},\textit{hi},\textit{ur},\textit{mr},\textit{pa},\textit{ta}, \textit{te}, \textit{ml}, \textit{kn}\} (on 0.1 Million tokens).}
	\label{fig21:mistral_set6_ppx_change}
\end{figure}
\begin{figure}[ht!]
	\centering
	\includegraphics[width=0.45\textwidth]{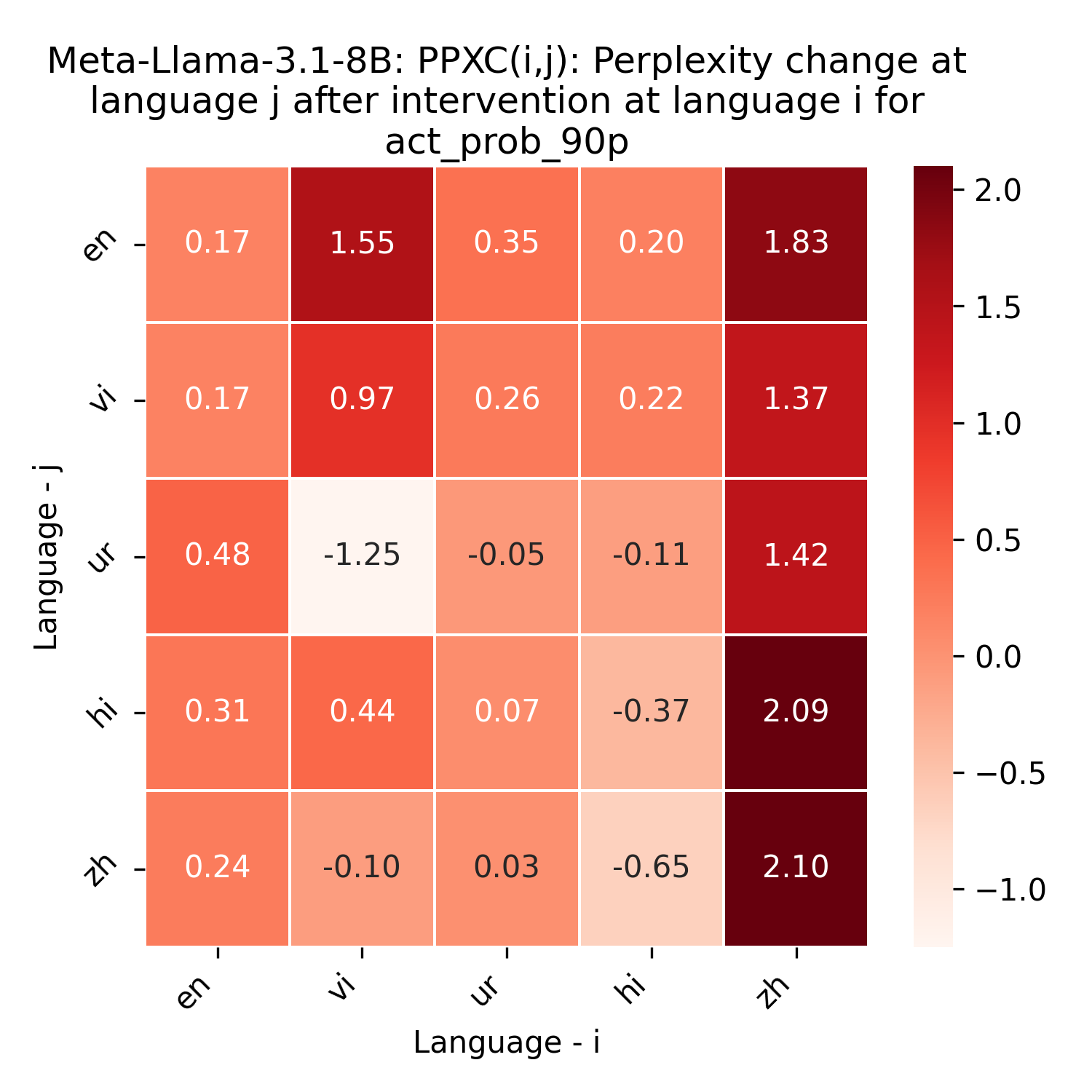}
	\caption{\footnotesize Llama 3.1: Perplexity change for Activation Probability 90p (on 0.1 Million tokens).}
	\label{fig22:llama3_act_ppx_change}
\end{figure}
\begin{figure}[ht!]
	\centering
	\includegraphics[width=0.45\textwidth]{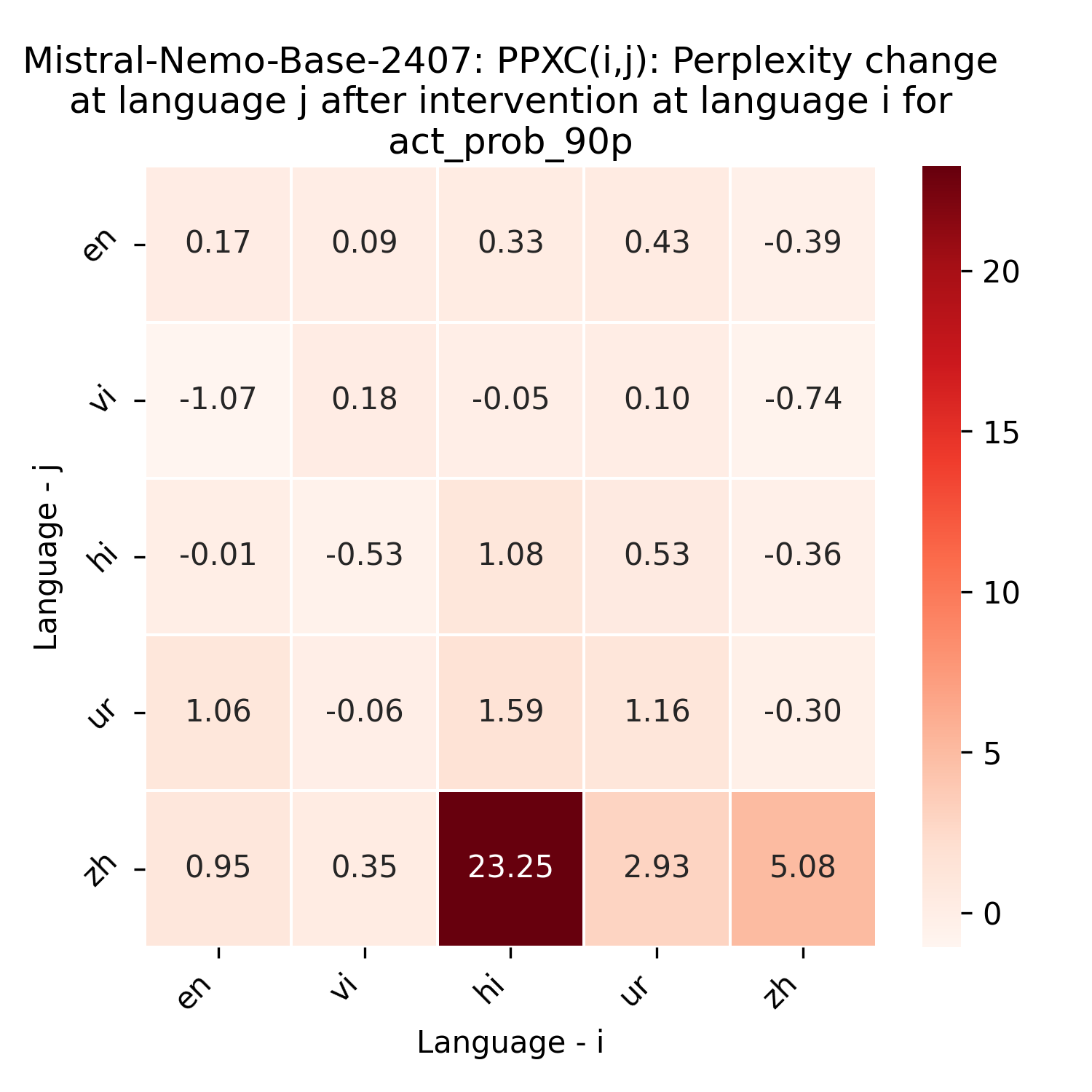}
	\caption{\footnotesize Mistral Nemo: Perplexity change for Activation Probability 90p (on 0.1 Million tokens).}
	\label{fig23:mistral_act_ppx_change}
\end{figure}

\clearpage

	\begin{table*}[ht!]
		\centering
		\renewcommand{\arraystretch}{1.13} 
		\setlength{\tabcolsep}{7pt} 
		\begin{small}
			\begin{tabular}{|c|c|c|c|c|c|c|c|c|c|c|}
				\hline
				\makecell{\textbf{Model \&} \\ \textbf{Method}}  & \makecell{\textbf{Eval} \\ \textbf{Lang}} & \textbf{No Int} & \textbf{Int} $\mathbf{\mu}$ & \textbf{Int P75} & \textbf{Int P90} & \textbf{Int P95} & \textbf{Int 0} & \textbf{Int P5} & \textbf{Int P10} & \textbf{Int P25} \\ 
				\hline
				\multirow{3}{*}{\makecell{Llama 3.1 \\ + \\ LAPE}} & vi & \textbf{80.5} & 79.5	& 79.1	& 79.0	& 78.9 & 79.8 & 73.6	& 77.7	& \textbf{80.5} \\ 
				& hi & 75.0 &	\textbf{75.2}	& 74.9	& 74.9 &	75.0		&  74.4 &	74.8	& 75.1	& 74.9\\ 
				& ur & 70.0 &	\textbf{70.4}	& 70.2	& 69.3	& 69.0	& 68.5 &		68.6	& 68.7	& 69.7 \\ 
				\hline
				\multirow{3}{*}{\makecell{Llama 3.1 \\ + \\ Act Prob 90p}} & vi & \textbf{80.5} &	78.2	& 78.9 &	79.3	& 79.5	& 79.0 &	76.5	& 77.4	& 78.0 \\ 
				& hi & \textbf{75.0}	& 74.1	& 72.6	& 71.8	& 71.2	& 73.7 &	75.0	& 74.6	& 74.2\\ 
				& ur & \textbf{70.0}	& 69.7	& 69.7	& 69.3	& 68.7	& 69.6&	69.4	& 69.5 &	69.5 \\ 
				\hline\hline
				\multirow{3}{*}{\makecell{Mistral Nemo \\ + \\ LAPE}} & vi & 80.5 &	80.4 &	80.5 &	\textbf{80.6} &	80.5 &	79.2 &	\textbf{80.6} &	80.5 &	80.3 \\ 
				& hi & \textbf{76.1} & 69.8 & 67.8 & 66.9 & 66.6 & 74.9 & 73.0 & 72.4 & 71.0\\ 
				& ur & 66.8 & 66.5 & 67.2 & 67.0 & 66.8 & \textbf{66.9} & 65.8 & 65.4 & 65.7\\ 
				\hline
				\multirow{3}{*}{\makecell{Mistral Nemo \\ + \\ Act Prob 90p}} & vi & 80.5 &	67.4 &	79.0 &	\textbf{81.1} &	\textbf{81.1} &	79.8 &	37.4 &	40.7 &	47.8 \\ 
				& hi & \textbf{76.1} &	72.2 &	73.7 &	74.5 &	74.4 &	74.5 &	62.4 &	66.3 &	69.3 \\ 
				& ur & \textbf{66.8} &	65.9 &	64.0 &	61.3 &	59.7 &	66.4 &	54.6 & 	61.6 &	65.0 \\ 
				\hline
			\end{tabular}
		\end{small}
		\caption{\footnotesize Full XNLI performance results, including additional statistical interventions. This table extends Table~\ref{tab:xnli} by incorporating multiple activation percentile-based interventions, including P75, P90, P95, P5, P10, and P25, alongside the mean and zero activation interventions.}

		\label{tab:xnli1}
	\end{table*}
	
	\begin{table*}[ht!]
		\centering
		\renewcommand{\arraystretch}{1.3} 
		\setlength{\tabcolsep}{3.5pt} 
		\begin{small}
			\begin{tabular}{|c|c|c|c|c|c|c|c|c|c|c|}
				\hline
				\makecell{\textbf{Model \&} \\ \textbf{Method}}  & \makecell{\textbf{Eval} \\ \textbf{Lang}} & \textbf{No Int} & \textbf{Int} $\mathbf{\mu}$ & \textbf{Int P75} & \textbf{Int P90} & \textbf{Int P95} & \textbf{Int 0} & \textbf{Int P5} & \textbf{Int P10} & \textbf{Int P25} \\ 
				\hline
				\multirow{3}{*}{\makecell{Llama 3.1 \\ + \\ LAPE}} & vi & 41 (73.5)	& 40 (72.9) &	36 (76.7)	& 31 (69.5) &	28 (66.8)	& 32  (69.2)	& 4 (35.8)	& 10 (43.2)	& 39 (71.5) \\ 
				& hi & 38 (64.1) & 40 (65.5) & 38 (66.4) & 36 (65.4) & 36 (66.3) & 23 (49.9) & 38 (62.8) & 37 (62.8) & 39 (65.3) \\ 
				& zh & 56 (77.5) & 10 (62.8) & 3 (58.1) & 3 (56.1) & 4 (52.9) & 33 (63.2) & 20 (49.1) & 33 (63.2) & 22 (67.8) \\ 
				\hline
				\multirow{3}{*}{\makecell{Llama 3.1 \\ + \\ Act Prob 90p}} & vi & 41 (73.6) & 39 (73.0) & 26 (65.8) & 23 (64.9) & 24 (64.8) & 42 (73.8) & 33 (67.8) & 36 (70.3) & 39 (72.0) \\ 
				& hi & 38 (64.1) & 34 (60.7) & 35 (62.3) & 36 (62.8) & 32 (61.9) & 38 (62.9) & 23 (54.9) & 31 (58.6) & 35 (60.7) \\ 
				& zh & 56 (77.5) & 61 (80.7) & 59 (79.5) & 56 (78.8) & 56 (78.4) & 55 (78.5) & 40 (61.7) & 50 (73.6) & 56 (79.3) \\ 
				\hline\hline
				\multirow{3}{*}{\makecell{Mistral Nemo \\ + \\ LAPE}} & vi & 39 (74.6) & 42 (76.8) & 40 (76.4) & 40 (75.0) & 39 (73.6) & 13 (45.0) & 10 (40.4) & 11 (41.2) & 37 (73.3)\\ 
				& hi & 38 (66.9) & 35 (65.9) & 38 (67.1) & 37 (66.6) & 34 (66.5) & 22 (51.7) & 36 (65.8) & 36 (66.1) & 36 (67.1)\\ 
				& zh & 47 (74.9) & 24 (74.0) & 4 (65.6) & 0 (61.6) & 0 (59.3) & 14 (53.3) & 1 (33.6) & 24 (68.9) & 35 (77.0)\\ 
				\hline
				\multirow{3}{*}{\makecell{Mistral Nemo \\ + \\ Act Prob 90p}} & vi & 39 (74.6) & 11 (43.3) & 26 (62.8) & 29 (63.9) & 15 (48.5) & 39 (74.5) & 0 (3.8) & 0 (6.5) & 2 (12.9)\\ 
				& hi & 38 (66.9) & 26 (54.4) & 34 (65.1) & 37 (68.9) & 37 (68.7) & 33 (63.8) &  0 (6.0) & 0 (11.9) & 12 (35.2)\\
				& zh & 47 (74.9) & 46 (77.4) & 26 (69.9) & 20 (59.8) & 15 (54.9) & 48 (76.2) & 0 (8.5) & 0 (17.0) & 9 (45.6)\\ 
				\hline
			\end{tabular}
		\end{small}
		\caption{\footnotesize Full XQuAD performance results, with exact match (EM) and F1 scores across various interventions. This table builds upon Table~\ref{tab:xquad}, expanding the analysis with additional intervention strategies. The results further validate the findings on test-time interventions and their impact on cross-lingual task performance.}
		\label{tab:xquad1}
	\end{table*}
	
	\begin{table*}[ht!]
		\centering
		\setlength{\tabcolsep}{7.5pt} 
		\begin{small}
			\begin{tabular}{|c|c|c|c|c|c|c|c|c|}
				\hline
				\textbf{Model} & \textbf{Lang} & \textbf{Act} $\mathbf{\mu}$ & \textbf{Act P75} & \textbf{Act P90} & \textbf{Act P95} & \textbf{Act P5} & \textbf{Act P10} & \textbf{Act P25} \\ 
				\hline
				\multirow{5}{*}{Llama 3.1} & en  & -0.2621 & -0.0493 & 0.1693 & 0.3106 & -0.7943 & -0.6801 & -0.4892 \\ 
				& vi & -0.2599 & -0.049 & 0.1648 & 0.3012 & -0.782 & -0.6709 & -0.4841 \\ 
				& hi & -0.2728 & -0.0648 & 0.144 & 0.2784 & -0.7893 & -0.6788 & -0.4934 \\ 
				& ur & -0.254 & -0.0498 & 0.1521 & 0.2804 & -0.7604 & -0.6511 & -0.4684 \\ 
				& zh & -0.2603 & -0.0505 & 0.1611 & 0.2958 & -0.7797 & -0.6689 & -0.4826 \\
				\hline
				\multirow{5}{*}{Mistral Nemo} & en & -0.3938 & -0.0798 & 0.2477 & 0.4629 & -1.1888 & -1.0149 & -0.7294 \\ 
				& vi & -0.399 & -0.0807 & 0.2454 & 0.4564 & -1.1975 & -1.0237 & -0.737 \\ 
				& hi & -0.4265 & -0.1147 & 0.2053 & 0.4133 & -1.2092 & -1.0392 & -0.7588 \\ 
				& ur & -0.397 & -0.0881 & 0.2252 & 0.4279 & -1.1722 & -1.0029 & -0.7238\\ 
				& zh & -0.3962 & -0.0715 & 0.2568 & 0.4678 & -1.2088 & -1.032 & -0.7388\\
				\hline
			\end{tabular}
		\end{small}
		\caption{\footnotesize Activation statistics for Llama 3.1 and Mistral Nemo across different languages for Wikipedia dataset. It provides an in-depth analysis of activation values, including mean activation values and key quantiles (P75, P90, P95, P5, P10, P25).}

		\label{tab:act}
	\end{table*}

    	\begin{table}[ht!]
		\centering
		 \renewcommand{\arraystretch}{1.15} 
		\setlength{\tabcolsep}{4.5pt} 
		\footnotesize
		\begin{tabular}{|c|c|c|c|c|c|c|}
			\hline
			\textbf{FTL} & \textbf{EL} & \textbf{No Int} & \textbf{Int} $\mathbf{\mu}$ & \textbf{Int P90} & \textbf{Int 0} & \textbf{Int P10} \\ 
			\hline
			\multicolumn{7}{|c|}{\textit{Llama 3.1 with LAPE}} \\ 
			\hline
			en & vi & \textbf{80.2} & 79.6 & 78.5 & 79.2 & 78.0 \\ 
			vi & vi & \textbf{80.1} & 79.5 & 78.6  & 79.2 & 78.0 \\  
			en+vi & vi & \textbf{80.1} & 79.4 & 78.5 & 79.1 & 78.0 \\
			\hline
			en & hi & \textbf{74.9} & 74.6 & 74.6 & 74.1  & 74.6 \\ 
			hi & hi & \textbf{74.9} & 74.6 & 74.5 &  74.3 & 74.6 \\  
			en+hi & hi & \textbf{74.9} & 74.5 & 74.5 &  74.3 & 74.7 \\
			\hline
			en & ur & 69.8 & \textbf{70.4} & 69.6 &  70.2 & 69.0 \\ 
			ur & ur & \textbf{69.8} & 70.5 & 69.5 & 70.4 & 69.1 \\  
			en+ur & ur & 70.0 & \textbf{70.6} & 69.5 & 70.3 & 68.9 \\
			\hline
			\multicolumn{7}{|c|}{\textit{Llama 3.1 with Act Prob 90p}} \\ 
			\hline
			en & vi & \textbf{80.3} & 78.0 & 79.1 & 78.5 & 77.5 \\ 
			vi & vi & \textbf{80.1} & 78.0 & 79.0 &  78.7 & 77.4 \\ 
			en+vi & vi & \textbf{80.1} & 78.0 & 78.9 & 78.8 & 77.4 \\
			\hline
			en & hi & \textbf{74.9} & 73.9 & 72.1 & 73.3 & 74.5 \\ 
			hi & hi & \textbf{74.9} & 73.8 & 72.1 & 73.9 & 74.6 \\ 
			en+hi & hi & \textbf{74.9} & 73.7 & 72.1 &  73.8& 74.6 \\
			\hline
			en & ur &69.9 & \textbf{70.1} & 69.4 & 70.0 & 69.5 \\ 
			ur & ur & 69.9 & 70.0 & 69.4 & \textbf{70.1} & 69.5 \\ 
			en+ur & ur & 69.8 & 69.9 & 69.3 & \textbf{70.1} & 69.6 \\
			\hline
		\end{tabular}
		\caption{\footnotesize Full language neuron fine-tuning results for XNLI. This table extends Table~\ref{tab:ft}, presenting fine-tuning experiments where language-specific neurons are updated in different configurations. It includes results for models fine-tuned on the source language alone, the target language alone, and both together, with evaluation of test-time interventions across multiple setups.}
		\label{tab:ft1}
	\end{table}

\begin{table}[ht!]
    \centering
    \renewcommand{\arraystretch}{1.3} 
    \setlength{\tabcolsep}{6pt} 
    \footnotesize
    \begin{tabular}{|c|c|c|c|}
        \hline
        \textbf{Model} & \textbf{FTL} &  \textbf{EL} & \textbf{No Int} \\ 
        \hline
        Llama 3.1 & rand & vi & 80.1 \\  
        Llama 3.1  & rand & hi & 74.8 \\  
        Llama 3.1  & rand & ur & 69.8 \\  
        \hline
        Mistral Nemo & rand  & vi & 80.4 \\  
        Mistral Nemo  & rand & vi & 75.6 \\  
        Mistral Nemo  & rand  & ur & 65.5 \\  
        \hline
    \end{tabular}
    \caption{\footnotesize Random neuron fine-tuning results for Llama 3.1 and Mistral Nemo. The table reports zero-shot performance without intervention (No Int) after fine-tuning randomly selected 10 neurons per layers instead of language-specific neurons.}
    \label{tab:rft}
\end{table}
\fi
\end{document}